%% file: main.tex
\documentclass[preprint,3p,times,twocolumn]{elsarticle}
\usepackage[hidelinks]{hyperref}

\journal{Robotics and Autonomous Systems,  
}
\usepackage{times}
\usepackage[table]{xcolor}
\definecolor{Light0}{rgb}{0.98, 0.95, 0.99}
\definecolor{Light1}{rgb}{0.98, 0.95, 0.90}
\definecolor{Light2}{rgb}{0.98, 0.98, 0.93}
\definecolor{Light3}{rgb}{0.98, 0.98, 1}

\usepackage[numbers]{natbib}
\usepackage{multicol}
\usepackage{graphics} 
\usepackage{pict2e}
\usepackage{dirtytalk}
\usepackage{epsfig} 
\usepackage{amsmath} 
\usepackage{amssymb}  
\usepackage{ upgreek }
\usepackage[linesnumbered,ruled]{algorithm2e}
\usepackage{tablefootnote}
\usepackage{xcolor}
\usepackage{graphicx}
\usepackage{float} 
\usepackage{threeparttable}
\usepackage{amsmath}
\usepackage{threeparttable, tablefootnote}

\floatstyle{ruled}
\newfloat{algorithm}{tbp}{loa}
\providecommand{\algorithmname}{Algorithm}
\floatname{algorithm}{\protect\algorithmname}
\usepackage{siunitx}
\usepackage{algpseudocode}
\usepackage{epstopdf}
\usepackage{multirow}
\usepackage[leftcaption]{sidecap}
\usepackage{booktabs}
\usepackage{amssymb}
\usepackage{arydshln}
\usepackage{lipsum}
\usepackage{comment}

\usepackage[nolist]{acronym} 
\input{acronyms.tex}

\input{macrosNew.tex}
\graphicspath{{figures/}}

\begin{document}

\begin{frontmatter}
	\title{
		Deep Model Predictive Variable Impedance Control
	}
	\author[ntnu]{Akhil S Anand\corref{mycorrespondingauthor}}
	\cortext[mycorrespondingauthor]{Corresponding author}
	\ead{akhil.s.anand@ntnu.no}
	
	\author[aalto]{Fares J.~Abu-Dakka}
	\ead{fares.abu-dakka@aalto.fi}
	
	\author[ntnu]{Jan Tommy Gravdahl}
	\ead{jan.tommy.gravdahl@ntnu.no}

	\address[ntnu]{Dept. of Engineering Cybernetics at Norwegian University of Science and Technology (NTNU), Trondheim, Norway.}
	\address[aalto]{Intelligent Robotics Group, Department of Electrical Engineering and Automation (EEA) at Aalto University, Aalto, Finland.}

	\begin{abstract}
	The capability to adapt compliance by varying muscle stiffness is crucial for dexterous manipulation skills in humans. Incorporating compliance in robot motor control is crucial to performing real-world force interaction tasks with human-level dexterity.  This work presents a Deep Model Predictive Variable Impedance Controller for compliant robotic manipulation which combines Variable Impedance Control with Model Predictive Control (MPC). A generalized Cartesian impedance model of a robot manipulator is learned using an exploration strategy maximizing the information gain. This model is used within an MPC framework to adapt the impedance parameters of a low-level variable impedance controller to achieve the desired compliance behavior for different manipulation tasks without any retraining or finetuning.  The deep Model Predictive Variable Impedance Control approach is evaluated using a Franka Emika Panda robotic manipulator operating on different manipulation tasks in simulations and real experiments. The proposed approach was compared with model-free and model-based reinforcement approaches in variable impedance control for transferability between tasks and performance. 
	\end{abstract}
	

\end{frontmatter}
\section{Introduction}

Manipulating objects is central to how humans interact with the real-world and even with a limitation of low-frequency biological feedback loops, we possess dexterous manipulation skills. Although the exact motor control mechanisms responsible for such skills remain unknown, the impedance modulation of the arm has been proposed as key mechanism \cite{bizzi1984posture, hogan1984organizing, kennedy2019stiffness}. While in robotics the feedback control loops can be operated at much higher frequencies, human-level dexterity is seldom achieved in real-world applications. Most real-world applications using robotic manipulators traditionally relied on trajectory planning and position control which is undesirable for dexterity, safety, energy efficiency, and constrained interactions. Human muscle actuators posses the impedance properties (stiffness and damping) \cite{hill1950series} which can be adapted by the neural control to achieve various manipulation behaviours.  
Motivated by human manipulation, \ac{ic} for robot control, introduced by Hogan in \cite{hogan1984impedance}, aims to couple the manipulator dynamics with its environment instead of treating it as an isolated system while designing control strategies. 
Unlike the more conventional control approaches, \ac{ic} attempts to implement a dynamic relation between manipulator variables such as end-point positions and forces rather than just control these variables independently. The use of \ac{ic} provides a feasible solution to overcome position uncertainties in order to avoid large impact forces since robots are controlled to modulate their motion or compliance according to force feedback \cite{khatib1987unified}. 

\ac{ic} is naturally extended to \ac{vic} where the impedance parameters are varied during the task \cite{ikeura1995variable, caldarelli2022perturbation}. \ac{vic} gained popularity in robotic research due the adaptability and safety properties. The emergence of learning algorithms has generated great interest in learning based \ac{vic}, where a learned policy is used to adapt the impedance gains in \ac{vilc} framework. In \cite{abu2020variable}, a detailed review of various learning approaches applied to \ac{vic} is presented. 
While using \ac{rl} for \ac{vilc}, or in general to robotics, the control policies obtained are task or scenario-specific, based on what they encountered during the learning. 
\begin{table*}[t]
	\centering
	\caption{{Comparison among state-of-the-art of \ac{vilc} approaches}}
	\begin{tabular}{m{0.15\linewidth}m{0.05\linewidth}m{0.05\linewidth}m{0.05\linewidth}m{0.05\linewidth}m{0.05\linewidth}m{0.05\linewidth}m{0.05\linewidth}m{0.15\linewidth}}
		\cline{2-9}
		& \multicolumn{1}{c}{\begin{tabular}[c]{@{}c@{}}Data-\\efficiency \end{tabular}} & \multicolumn{1}{c}{\begin{tabular}[c]{@{}c@{}}Task\\ transferability \end{tabular}} & \multicolumn{1}{c}{\begin{tabular}[c]{@{}c@{}}Model-based/\\  Model-free \end{tabular} } & \multicolumn{1}{c}{\begin{tabular}[c]{@{}c@{}}Computation\\  time \end{tabular} } & \multicolumn{1}{c}{\begin{tabular}[c]{@{}c@{}}force-/position-\\  based \ac{vic} \end{tabular} }\\ \hline
		\cite{martin2019variable, beltran2020variable, bogdanovic2020learning,varin2019comparison} 		& \multicolumn{1}{c}{low} & \multicolumn{1}{c}{-} & \multicolumn{1}{c}{model-free} & \multicolumn{1}{c}{low} & \multicolumn{1}{c}{force} \\ \hline
		\cite{beltran2020learning,kim2022scape} 		& \multicolumn{1}{c}{low} & \multicolumn{1}{c}{-} & \multicolumn{1}{c}{model-free} & \multicolumn{1}{c}{low} & \multicolumn{1}{c}{position} \\ \hline
		\cite{buchli2011learning} 	& \multicolumn{1}{c}{high} & \multicolumn{1}{c}{-} & \multicolumn{1}{c}{model-free} & \multicolumn{1}{c}{low} & \multicolumn{1}{c}{force} \\ \hline
		\cite{li2018efficient, roveda2020model}		& \multicolumn{1}{c}{high} & \multicolumn{1}{c}{-} & \multicolumn{1}{c}{model-based} & \multicolumn{1}{c}{high} & \multicolumn{1}{c}{position}\\ \hline
		\cite{anand2022evaluation}	& \multicolumn{1}{c}{high} & \multicolumn{1}{c}{-} & \multicolumn{1}{c}{model-based} & \multicolumn{1}{c}{high} & \multicolumn{1}{c}{force}\\ \hline
		Our \acs{mpvic}	& \multicolumn{1}{c}{high} & \multicolumn{1}{c}{\checkmark} & \multicolumn{1}{c}{model-based} & \multicolumn{1}{c}{high} & \multicolumn{1}{c}{force} \\ \hline
	\end{tabular}
	\label{tab:statOfArt}
\end{table*}
\ac{mpc} offers a framework to formulate the control systems as an optimization problem based on a  system model and an optimization objective \cite{camacho2013model}. \ac{mpc} approaches are widely used in robotic control when a model of the system dynamics is available. 
In the case of a robot controlled using a Cartesian space \ac{vic}, it is possible to learn a Cartesian impedance model of the robot and utilize it in an \ac{mpc} framework to optimize control polices over various tasks by providing suitable optimization objectives. While complex dynamics can be estimated from data using \ac{nn}s, they suffer from over-fitting and can not quantify uncertainties. \ac{gp} models are capable of modelling the uncertainty, but they don't scale well with high-dimensional data. \ac{penn} models introduced in \cite{chua2018deep} overcome these limitations of \ac{nn}s, offering a way to quantify both aleatoric and epistemic uncertainties. 

In this paper, we propose a deep \ac{mpvic} framework, where a \ac{nn} based Cartesian impedance model of the robotic manipulator is used in a \ac{cem}-based \ac{mpc} for online adaptation of the impedance parameters of a \ac{vic}. This deep \ac{mpvic} framework is utilized to learn impedance adaptation strategy for various robotic manipulation tasks by specifying a suitable cost function. {The main contributions of this paper are:  

\begin{itemize}
    \item a novel \ac{vic} framework, we call it deep \ac{mpvic}, which combines a \ac{cem}-based \ac{mpc} with \ac{penn} dynamical model for compliant robotic manipulation.
    \item the deep \ac{mpvic} framework learns a generalized Cartesian impedance model of the robot to facilitate the transferability between completely different manipulation tasks without any need of relearning the model. 
    \item an uncertainty-based exploration scheme is integrated into the proposed framework to facilitate learning a generalized model efficiently from fewer samples. 
    \item an extensive evaluation in simulation and real setups, in addition to a comparison between our approach and the state-of-the art model-free  and model-based \ac{rl} approaches on transferrability and performance.
\end{itemize}       

}

The rest of the paper is organized as follows. {\secref{related_works} describes the existing references relevant to our work.} \secref{background} briefly introduces the necessary background knowledge, in \secref{mpic_frmework} presents the details of the deep \ac{mpvic} framework proposed. \secref{sec:evaluation} presents the evaluation of our approach on simulation and experimental setups using Franka Panda robotic manipulator. Detailed discussion on the results and the limitations of our approach is presented in \secref{disussions} and conclusion in \secref{conclusions}.

\section{Related Work} \label{related_works}

Recently, \ac{rl} has been explored largely for \ac{vilc} research. However, \ac{rl} demands large amount of data samples/interactions to obtain high performance. {References \cite{martin2019variable, beltran2020variable, beltran2020learning, bogdanovic2020learning, varin2019comparison} are some examples of using deep \ac{rl} for \ac{vilc} applied to different robotic manipulation tasks.} All these approaches could learn complex \ac{vic} policies for specific tasks, however, at the expense of sample efficiency. {  Ref. \cite{kim2022scape} combines human demonstrations with \ac{rl}, providing improved sample efficiency for learning stiffness control policies. But it is not suitable for force-based \ac{vic}, as unlike stiffness values the impedance parameters can not be estimated directly from kinesthetic demonstrations used in \cite{kim2022scape}.} Ref. \cite{buchli2011learning} demonstrated model-free \ac{rl} based \ac{vilc} using \ac{dmp} policy and \ac{pi2}, which is sample efficient but  fails to scale to complex policies. Whereas our \ac{mpc} based approach is scalable to complex problems with a \ac{nn} dynamics model. {Apart from sample efficiency a major drawback of the referenced \ac{rl} based approaches is their inability to easy transfer of a learned policy to a different task. In practice,  retraining the policy is necessary, which is difficult in real-world robotic tasks. In contrast, our deep-\ac{mpvic} framework use a generalized Cartesian impedance model of the robot with an \ac{mpc} policy that can be used for multiple tasks by designing suitable cost functions.}

Alternatively, \ac{mbrl} approaches offer a sample efficient framework leveraging on the model. In \cite{li2018efficient} \ac{mbrl} is used for learning position-based \ac{vic} on industrial robots using \ac{gp} models.  Ref. \cite{anand2022evaluation} used a similar approach for force-based \ac{vic} for contact sensitive tasks. {Both of these approaches utilizes \ac{gp} models and PILCO algorithm limiting its use to less complex tasks with smooth dynamics and relatively simple policies and reward structure.}   In \cite{roveda2020model},  a \ac{pets} approach is utilized for learning position-based \ac{vic} strategy for \ac{hrc} tasks. Similar to \ac{rl} approaches referenced earlier, all of these \ac{mbrl}-based approaches are task specific and  generally lacks the performance of model-free \ac{rl} approaches \cite{wang2019benchmarking}. Unlike aforementioned \ac{vilc} approaches, our deep \ac{mpvic} is not only able to adapt to new situations of the same task, but also it is transferable to new tasks using the same trained model without any need to re-train or train a new model. Transferability between tasks is achieved by combining a generalised Cartesian impedance model with an \ac{mpc} scheme. A comparison between existing \ac{rl}-based \ac{vilc} approaches is summarised in Table \ref{tab:statOfArt}. 

In literature, \ac{mpc} is used in robotic interaction control for manipulations tasks  \cite{minniti2021model, gold2022model}, where \ac{mpc} optimizes the robot control input but not the stiffness itself, while in our approach the \ac{mpc} adapt the stiffness values directly. It is possible to couple our deep \ac{mpvic} with the approach in \cite{minniti2021model} where it can be used as a low level optimizer to solve additional constraints. Haninger\etal~\cite{haninger2022model} used an \ac{mpc} scheme with \ac{gp} models for human-robot interaction tasks. The \ac{mpc} scheme used could optimize the impedance parameters for an admittance controller, but it is task specific as the human force model is estimated from demonstrations as a function of robot states. Using \ac{gp} models limits the complexity and generalizability of the model as pointed out by the authors in \cite{haninger2022model}. Unlike \cite{haninger2022model}, we optimize the impedance parameters for a force-based \ac{vic} in our deep \ac{mpvic} framework using \ac{penn} to model the Cartesian impedance behaviour of the robot manipulator.

For efficient model learning in terms of sample efficiency, uncertainty-based exploration with ensembles of \ac{nn}s  has been proposed in prior works \cite{shyam2019model, sekar2020planning, yao2021sample, pathak2019self}. The basis for uncertainty-based exploration for model learning is derived from the expected information gain formulation in \cite{lindley1956measure}.  In \cite{pathak2017curiosity} this approach is termed  \textit{curiosity-driven exploration}. The model uncertainty is evaluated based on the variance of the model in predicting the next state. {We incorporated \textit{curiosity-driven exploration} to our deep \ac{mpvic} framework to learn a generalized Cartesian impedance model sample efficiently.}


\section{Background}\label{background}
\subsection{Robot Manipulator Dynamics}
For a rigid $n$-DOF robotic arm, the task space formulation of the robot dynamics is given by
\begin{equation}
    {\bm{\Lambda}}(\bm{q})\bm{\ddot{x}} + \bm{\Gamma}(\bm{q},\bm{\Dot{q}}) \bm{\dot{x}} + \bm{\eta}(\bm{q}) = \bm{f_c} - \bm{f_{ext}} \, ,
    \label{task_space}
\end{equation}
where $\bm{\dot{x}}, \bm{\ddot{x}}$ are velocity and acceleration of the robot end-effector in task space, $\bm{f_c}$ is the task space control force, $\bm{f_{ext}}$ is the external force, $\bm{\Gamma}(\bm{q},\bm{\Dot{q}}) \in \bm{\mathbb{R}^{6 \times 6}}$ is a matrix representing the centrifugal and Coriolis effects, and $\bm{\eta}(\bm{q}) = \bm{J^{-T} {g}}(\bm{q})\in \bm{\mathbb{R}^{6 \times 1}}$ is the gravitational force, where $\bm{g}(\bm{q})$ is the joint space forces and torques. The Cartesian inertia matrix is denoted as $\bm{\Lambda}(\bm{q}) = (\bm{J} \bm{H}(\bm{q})^{\bm{-1}}\bm{J^T})\bm{^{-1}} \in \bm{\mathbb{R}^{6 \times 6}}$,
 where $\bm{H}(\bm{q}) \in \mathbb{R}^{n \times n}$ is the  joint space inertia matrix and $\bm{J}$ is the end-effector geometric Jacobian. By additionally knowing the joint space centrifugal and Coriolis matrix, $\bm{V}(\bm{q},\bm{\Dot{q}})$, the corresponding task space matrix is given by, 
 \begin{equation}
      \bm{\Gamma}(\bm{q},\bm{\Dot{q}}) = \bm{{J}}^{-T}\bm{V}(\bm{q},\bm{\Dot{q}})\bm{J}^{-1}-\bm{\Lambda}(\bm{q})\bm{\Dot{J}}\bm{J}^{-1} \, .
 \end{equation}

\subsection{Variable Impedance Control}
\ac{vic} is designed to achieve force regulation by adjusting the system impedance \cite{huang1992compliant}, via the adaptation of the inertia, damping, and stiffness components.  In the presence of a force and torque sensor measuring $\bm{f_{ext}}$, impedance control can be implemented by enabling inertia shaping \cite{villani2016force}. Casting the control law
\begin{equation}
    \bm{{f}_c} = \bm{\Lambda}(\bm{q})\bm{\alpha} + \bm{\Gamma}(\bm{q},\bm{\Dot{q}}) \bm{\Dot{x}} + \bm{\eta}(\bm{q}) + \bm{f_{ext}} \, ,
    \label{h_c}
\end{equation}
into the dynamic model in (\ref{task_space}) results in $\bm{\ddot{x}} = \bm{\alpha}$, $\bm{\alpha}$ being the control input denoting acceleration with respect to the base frame. 
In task space \ac{ic}, the objective is to maintain a dynamics relationship \eqref{closed-loop_inertia-shaping} between the external force, $\bm{f_{ext}}$, and the error in position $\bm{\delta{x}}=\bm{x^r} - \bm{x}$, velocity  $\delta{\bm{\dot{x}}}= \bm{\dot{x}^r} - \bm{\dot{x}}$ and acceleration $\delta{\bm{\ddot{x}}} = \bm{\ddot{x}^r} - \bm{\ddot{x}}$. This dynamic relationship that governs the interaction is modeled as a mass-spring-damper system as follows
\begin{equation}
    \bm{M} \delta \bm{\ddot{x}} + \bm{D} \delta \bm{\dot{x}} + \bm{K} \delta \bm{x} = \bm{{f}_{ext}} \, ,
    \label{closed-loop_inertia-shaping}
\end{equation}
where $\bm{M}$, $\bm{D}$ and $\bm{K}$ are \ac{spd} matrices, adjustable impedance parameters, representing inertia, damping and stiffness terms, respectively. This desired dynamic behaviour \eqref{closed-loop_inertia-shaping} can be achieved using the following control law,
\begin{equation}
    \bm{\alpha} = \bm{\ddot{x}} + \bm{M^{-1}}(\bm{D} \delta \bm{\dot{x}} + \bm{K} \delta \bm{x} - \bm{f_{ext}}) \, .
    \label{alpha^e}
\end{equation}

With no external force acting on the manipulator, under this control scheme, the end-effector will asymptotically follow the desired trajectory. In the presence of external forces, the compliant behavior of the end-effector is described by (\ref{closed-loop_inertia-shaping}).

\begin{figure*}[t]
	\centering
	\def\svgwidth{1\linewidth}
	{\fontsize{6}{6}
		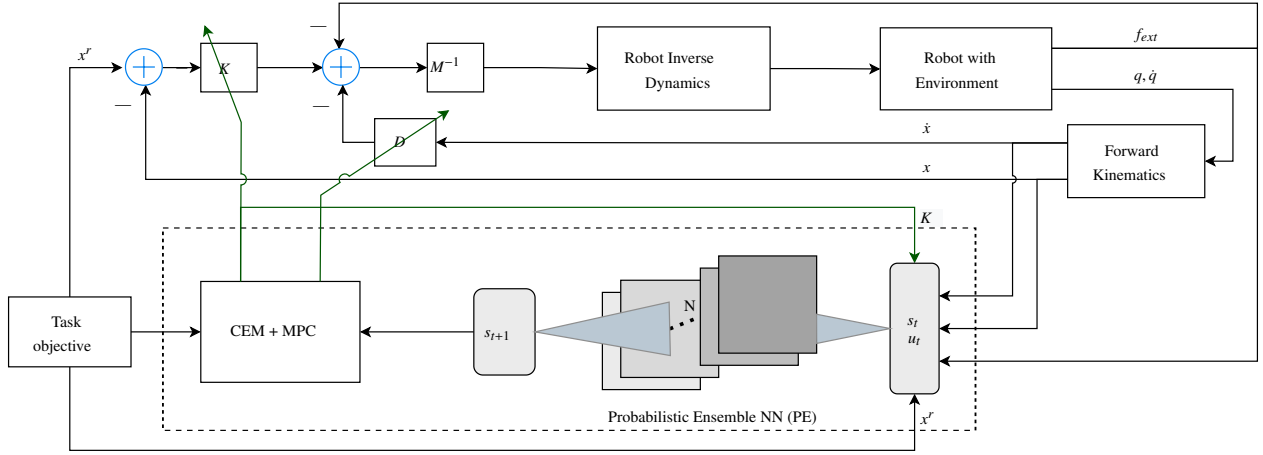}
	\caption{Block diagram of the deep \ac{mpvic} with \ac{penn} Cartesian impedance model.}
	\label{fig:framework}
\end{figure*}

\subsection{\acf{penn}}
\ac{penn} \cite{chua2018deep} is a \ac{nn} based model approach capable of learning uncertainty-aware \ac{nn} dynamics models including both aleatoric and epistemic uncertainties. The output neurons of the probabilistic \ac{nn} parameterize a probability distribution function, which can capture the aleatoric uncertainty of the model. Using multiple such networks in an ensemble can capture the epistemic uncertainty. The predictive \ac{penn} model trained with negative log prediction probability as a loss function can be defined to output a Gaussian distribution with diagonal covariance \cite{chua2018deep}.

\subsection{\ac{cem} based \ac{mpc}}
The \ac{cem} \cite{botev2013cross} offers a gradient free optimization scheme, coupling it with an \ac{mpc} allows us to optimize an action sequence using the learned model. \ac{cem} samples multiple action sequences from a time-evolving distribution which is usually modeled as a Gaussian distribution $u_{t: t+H} \sim \mathcal{N}\left(\mu_{t: t+H}, \operatorname{diag}\left(\sigma_{t: t+H}^{2}\right)\right)$, where these action sequences are evaluated on the learned dynamical model with respect to a cost function. The sampling distribution,  $\mu_{t: t+H}, \sigma_{t: t+H}^{2}$ is then updated based on best $\mathcal{N}$ trajectories. Safety can be directly incorporated into \ac{cem}-based optimization by sorting the samples based on constraint satisfaction values \cite{wen2018constrained}.

\section{Deep \acf{mpvic} Framework} \label{mpic_frmework}

The deep \ac{mpvic} framework is formulated to optimize a \ac{vic} utilizing a learned \ac{penn} based Cartesian impedance model of the robot manipulator within a \ac{cem} based \ac{mpc}. 

\subsection{Learning Cartesian Impedance Model }
A Cartesian impedance model of the robot manipulator system controlled using a \ac{vic} is learned as a \ac{penn} { model in an \ac{mbrl} setting alternating between model learning and \ac{cem} based exploration strategy.} To learn a generalized model, an exploration strategy is designed to minimize the epistemic uncertainty of the model across the entire state space. The exploration strategy chooses the actions which maximized the epistemic uncertainty estimate from \ac{penn}. Given a \ac{penn} model $\tilde f$ of $B$ bootstrap models $\tilde f_b$,  the uncertainty of the model prediction at current state can be estimated by calculating the model variance, $\rho = \sigma^2$, given by
\begin{equation} \label{eq:uncertainty}
    \rho (s,u) =  \frac{1}{B-1} \sum_{b=1}^B \left(\tilde f_b(s,u) - \overline{\tilde f(s,u)} \right) ^2 \,.
\end{equation}

The designed exploration strategy will excite the system in areas in its state space where the model is more uncertain, thereby maximizing the information gain during exploration. This exploration strategy enables learning a generalized model in a sample-efficient way. A \ac{cem}-based \ac{mpc} is used for exploration while the cost is defined to maximize the variance of the outputs from all the individual \ac{nn} models in the \ac{penn}, $C =  \sigma^2(s_t, u_t)$. {The model learning approach is summarized in Algorithm \ref{alg:model_learning}.} Learning a model with low epistemic uncertainty over the entire state-space facilitates reusing the model for different tasks. 

A free-space unconstrained manipulation task where the robot has to interact with its external environment can be described by a scenario where a robot in its current state $s_{t}$ under the influence of an external force or sensed force $f_{t}$ provided with a goal state $s_{t}^{r}$ and a control input $u_{t}$ transitions to the next state $s_{t+1}$. For a robot manipulator controlled by a \ac{vic} the stiffness matrix $\bm{K}$ can be considered as the control input $u_{t}$, where the damping matrix $ \bm{D}= 2\sqrt{\bm{K}}$. The dynamics model shown in  \figref{fig:framework} represents a generalized Cartesian behaviour of a unconstrained end-effector of a robot manipulator controlled by a \ac{vic}. 

\begin{algorithm}
\caption{Learning a generalized Cartesian impedance model}
\label{alg:model_learning}

Initialize dynamics model $\tilde f$\,.\\
Populate dataset $\mathcal{D}$ using random controller for $n$ initial trials\,.

\For{ $k \gets 1$ \KwTo $K$ Trials}{
    Train dynamics model $\tilde f$ on $\mathcal{D}$\,.\;\\
    \For{$t \gets 1$ \KwTo TaskHorizon}{
        \For{Actions $u_{t:t+T} \sim$\ac{cem}$(\cdot)$, 1 \KwTo \ac{cem} Iterations}{
            Evaluate and sort the actions by maximizing the uncertainty estimate in \eqref{eq:uncertainty}\,.
        }
        Execute first action $u_t^*$ from optimal action sequence $u_{t:t+T}^*$\,.\;\\
        Record outcome: $\mathcal{D} \gets \mathcal{D} \cup (s_t, u_t, s_{t+1})$\,.
    }
}
\end{algorithm}

\subsection{Impedance Adaptation}
\begin{algorithm}
\caption{deep \ac{mpvic}}
\label{alg:deep-mpic}
Given a cost function $C$ and a \ac{penn} dynamics model $\tilde f$\,.\\
    \textbf{\ac{mpc} based optimization}\\
    \For{$t \gets 1$ \KwTo TaskHorizon}{
        \textbf{\ac{cem}-based optimization} \\
        \For {$i \gets 1$ \KwTo \ac{cem} Iterations}{
            \textbf{Generate N samples}\,.\\
            Sample $N$ stiffness profiles $K_{t:t+T} \sim$ \ac{cem}$(\cdot)$\,. \\
            \textbf{Evaluate samples}\,.\\
            Calculate $C$ for all $K_{t:t+T}$ on $\tilde f$\ with actions $[K_{t:t+T}, f_{t}, s_{t}^{r}]$ using trajectory sampling \cite{chua2018deep}\,. \\
            Sort stiffness profiles $K $ based on $C$. \\
            Update CEM(·) distribution\,.\\
            Choose optimal $K^*$ where $C$ is minimum\,.
        }
        \textbf{Adapt the impedance parameters of \ac{vic}}\,.\\
        Execute first action $K_t^*$ from optimal action sequence $K_{t:t+T}^*$\,.\\
    }
\end{algorithm}

The compliant behavior of the robot end-effector can be optimized by designing a suitable impedance adaptation strategy. The Cartesian impedance model of the robotic system can be utilized in a \ac{mpc} framework to adapt the impedance parameters of the \ac{vic} by designing a suitable optimization objective as shown in \figref{fig:framework}. At every time-step, an \ac{mpc} with a horizon length of $n$, samples the current state and optimizes a control trajectory $u_{t : t+n}$ for $n$ future time-steps and applies the first control input, $u_t$, to the system. A gradient-free optimization method, \ac{cem} is used in an \ac{mpc} setting to optimize the controller over the \ac{penn} model. The proposed deep \ac{mpvic} approach utilizing \ac{penn} models is described in Algorithm \ref{alg:deep-mpic}. The objective of the impedance adaptation strategy is to achieve the manipulation task requirement while executing a desired level of compliance. A cost function describing the task objective and the compliance objective is designed for the \ac{cem}-based \ac{mpc} as,
\begin{equation}\label{eq:cost_function}
C\left(s_{t}, u_{t}\right)= \delta{s_{t}^{T}} \mathbf{Q}_{t} \delta{s_{t}} +\lambda (K_{t})^{T} \mathbf{R}_{t} \lambda(K_{t}) \, ,
\end{equation}
where $\lambda(K_{t})$ are the eigenvalues of the stiffness matrix represented in a vector form,  $\delta{s_{t}} = s_{t}^{r} - s_{t} $ and $\mathbf{Q}_{t}$ and $\mathbf{R}_{t}$ are diagonal gain matrices for task and compliance components respectively. These gain matrices can be either constant or can be a function of the robots states. The \ac{mpc} output behaviour will be tightly coupled with the gain matrices. In case of reference tracking tasks we chose $\mathbf{Q}_{t}$ to be a linear function of $\norm{\delta{s_{t}}}$ so that \ac{mpc} will penalize larger deviations from target more than small deviations. 

\begin{figure*}[t]
	\centering
	\def\svgwidth{1\linewidth}
	{\fontsize{8}{8}
		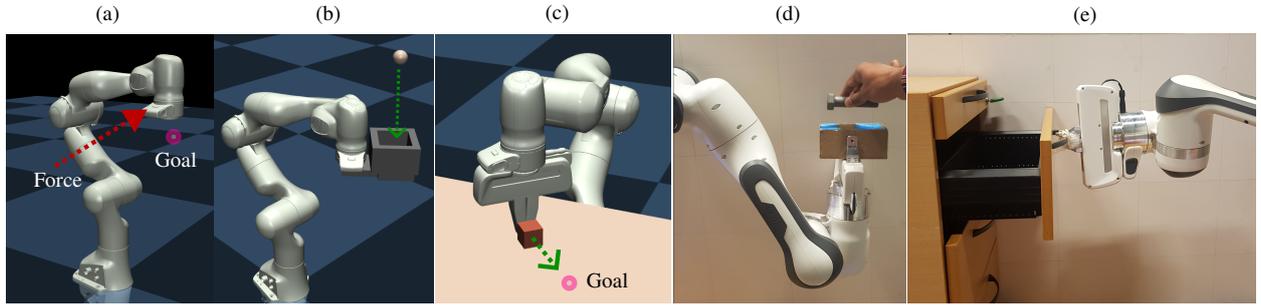}
	\caption{Three simulation tasks, (a) Cartesian compliance task: the robot manipulator end-effector should hold its pose in the Cartesian space compliantly while reacting to the external forces acting on it. (b) Reacting to falling object: The robot manipulator with cup end-effector should hold a Cartesian position while smoothly catching a ball of weight \num{0.5}\SI{}{\gram} falling into the cup. (c) Pushing task: A robot manipulator with a gripper end-effector should push an object over a rigid surface with friction to a target position. Two experimental tasks, (d) Reacting to falling objects:  robot end-effector is fitted with a tray, where objects of different weights are dropped into the tray at regular intervals. (e) Drawer opening task: Robot manipulator opening a table drawer. }
	\label{fig:simulation_tasks}
\end{figure*}
\section{Experiments and Evaluation} \label{sec:evaluation}
For evaluation we consider only the stiffness adaptation along the position of the robot manipulator while keeping the stiffness values along orientations constant. However, before evaluation, we first need to learn the Cartesian impedance model of the robot manipulator. To do so, a free-space goal reaching task with random external force is used to train \ac{penn} model with ensembles of $5$ \ac{nn}s with $3$ hidden layers, each with $256$ neurons. Its state space is chosen as $s = [x, y, z, \dot{x}, \dot{y}, \dot{z}]$,  while the sensed external forces are denoted as $f=[f_{ext}^{x}, f_{ext}^{y}, f_{ext}^{z}]$. $s^{r}=[x^{r}, y^{r}, z^{r}]$ represents the target positions in  $x,y$ and $z$ directions, $\bm{K}$ denotes the Cartesian stiffness matrix. The damping matrix is chosen as $ \bm{D} = 2 \sqrt{\bm{K}}$. \ac{cem} is used to optimize the exploration strategy based on uncertainty maximization. The control frequency for low-level \ac{vic} is set at \num{100}\SI{}{\hertz}. For learning the model, the robot manipulator is excited at every time-step with random $f_{ext} \in {[-20,20]}$ \SI{}{\newton}  and $s_{t}^{r}$, where  $x_{t}^{r}, y_{t}^{r} , z_{t}^{r}  \in {[-10,10]}$ \SI{}{\cm}. The gain matrices $\mathbf{Q}$ and $\mathbf{R}$ are kept constants for a specific task. However, while transferring to a new task, they can be scaled using a scalar values $\alpha_{Q}$ and $\alpha_{R}$ as $\mathbf{Q_{new}} = \mathbf{Q} * \alpha_{Q}$ and $\mathbf{R_{new}} = \mathbf{R} * \alpha_{R}$ respectively to trade off between compliance and accuracy depending on the task requirement. 
The model was trained for \num{100000} time-steps with a control-frequency of \num{10}\SI{}{\hertz} which is equivalent to \num{2.77} \SI{}{\hour} of real-world training. For experiments, a prior model  estimated in simulations over \num{50000} time-steps is fine-tuned in the experimental scenario instead of learning from scratch. The model was fine-tuned for \num{10000} time-steps which is equivalent to \num{33.33} \SI{}{\minute} of real-world training. Similar to in simulations random external forces were manually applied to the robot end-effector.

After learning the Cartesian impedance model of the manipulator, and to evaluate the effectiveness of the proposed deep \ac{mpvic}, three different simulation tasks and two experimental tasks using a Franka Emika Panda manipulator are designed. The tasks demanding real time stiffness adaptation are chosen for evaluating the stiffness profile generated by the deep \ac{mpvic} controller. The three different simulation tasks are modeled in the MuJoCo physics simulation framework \cite{todorov2012mujoco}, see \figref{fig:simulation_tasks} (a), (b) and (c). The two real experimental scenarios are shown in \figref{fig:simulation_tasks} (d) and (e). 
\textit{In simulations}, the population size for \ac{cem} is chosen as \num{200} and elite size  of \num{40} and learning rate of \num{0.1} and number of \ac{cem} iterations as \num{10}. The \ac{mpc} planning horizon is set as \num{5}.   
\textit{While for the real experiments}, the control frequency is set as \num{5}\SI{}{\hertz}.  The \ac{cem} is chosen as \num{64} and elite size  of \num{32} and learning rate of \num{0.5}, number of \ac{cem} iterations as \num{5} and \ac{mpc} planning horizon is set as \num{5}. {In all the simulations and experiments the model described here is used without any further fine-tuning. Here we consider only fixed goal states, therefore $s_{t}^{r}$ is a constant value, $s^r$ for all timesteps. }

\subsection{Simulations}
\begin{figure*}[t]
	\centering
	\def\svgwidth{1\linewidth}
	{\fontsize{8}{8}
		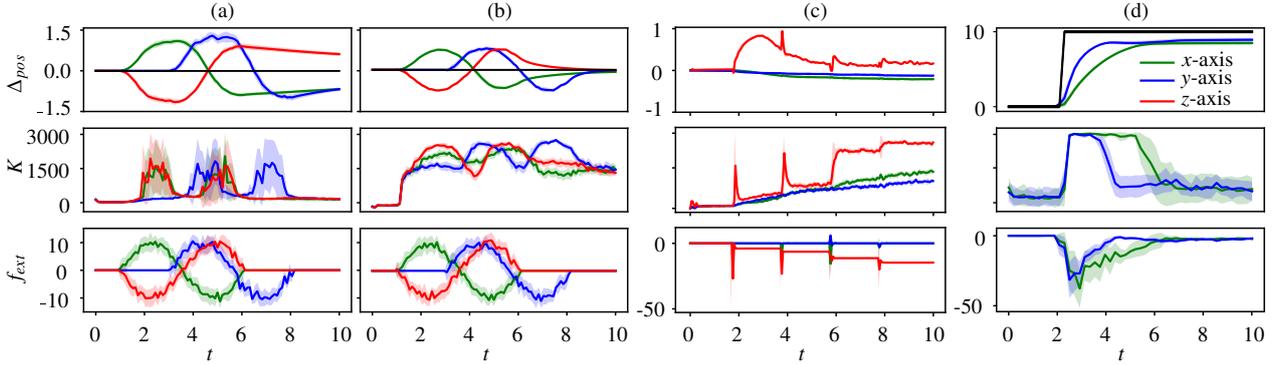}\vspace{-.5cm}
	\caption{Simulations: (a) and (b), (Cartesian compliance behaviour), results from \num{20} trials where a sinusoidal force profile with amplitude of \num{10} \SI{}{\newton} with a random noise of $(\pm{5})$ \SI{}{\newton} is applied to the robot end-effector. (a) High compliant behaviour optimized using a cost function with larger compliance factor $\alpha_{R} = 0.1$, (b) Low compliant behaviour optimized using a cost function with $\alpha_{R} = 0.01$. (c), (Reacting to falling objects) The robot is initialized at a rest position being very compliant with $K \to 0$. Objects of different weights are dropped at regular intervals of $2$\SI{}{\second}, from random heights between $(0.5 -1.0)$ \SI{}{\metre}. Results shown are here are over \num{10} such random trials with $\alpha_{R} = 0.1$. (d), (Pushing task) Robot with a gripper end-effector is at rest with $K \to 0$.  At $t=1$ \SI{}{\second}, it is commanded to push an object to a target position given by $\Delta_{pos}$ of \num{10} \SI{}{\cm} in $x \text{and y}$ direction on a surface. The results shown here are over 10 trials with objects of random weights between $(0.5-3.0)$
	\SI{}{\kilogram} and $\alpha_{R} = 0.1$.}
	\label{fig:sims}
\end{figure*}

\textbf{Cartesian compliant behavior :} In this task (\figref{fig:simulation_tasks} (a)), the robot is expected to behave highly compliant to hold its pose thereby spending minimum energy at rest. Upon applying an external force to the robot-end-effector it is expected to counter the force by adapting its stiffness such that it achieves a new rest position close to the initial position. This task is ideal to test the impedance adaptation strategy as it needs to increase the stiffness in case of large external forces and larger deviation from its initial position. Two scenarios with different compliance behavior are evaluated here by changing the compliance maximization component in the cost function. The results in \figref{fig:sims} (a) and (b) show that the robot which is highly compliant at rest adapts the stiffness in response to the external forces and deviation from the rest position. Having a higher value of compliance factor $\alpha_{R}$ allows for larger deviations from the initial position when applied with an external force while having a lower $\alpha_{R}$ limits this deviation. It is also noted that higher $\alpha_{R}$ results in noisy stiffness adaption behavior as larger $\Delta_{pos}$ (the deviation from the desired pose) creates larger gradients in the cost function.

\textbf{Reacting to falling object:}
In this task (\figref{fig:simulation_tasks} (b)), a robot with a cup end-effector that is highly compliant at rest position is expected to react optimally to objects falling into the cup end-effector. Four different objects are dropped from different heights to the cup in different trials resulting in large variations in the impact force.  The desired behavior of the robot is not to deviate largely from the rest position while reacting to the falling objects while not being very stiff. The resulting robot behavior is shown in \figref{fig:sims} (c), which shows a sudden increase in $K_{z}$ upon a spike in $f_{ext}$ in $z$ direction induced by the impact of the falling object. The robot increases its stiffness every time a new object is falling to the cup and maintains a higher level of stiffness during the later phases to hold the robot back to a new rest position.

\textbf{Pushing task:}
In this task (\figref{fig:simulation_tasks} (c)), the robot is expected to push a cube-shaped object to a target position on a surface with friction. Here, $K_{z}$ is set constant as \num{1000} as the robot is not expected to move in $z$ direction. Stiffness in $x$ and $y$ directions are optimized to push the object to the target while being compliant and stiff only when necessary. The results in \figref{fig:sims} (d) show that the stiffness is increased to its upper limit in the pushing directions initially to overcome the static friction. Upon reaching close to the target position the stiffness is decreased to be more compliant. 

{
\subsection{Comparison with Model-free/based \texorpdfstring{\ac{rl}}{}:}
The deep \ac{mpvic} is compared with \ac{rl} based \ac{vilc} approaches for their transferability between tasks which is the main contribution of this work while also comparing their performance. Specifically, in these comparisons, we utilize the \ac{penn} model trained with curiosity driven exploration with our deep \ac{mpvic} for different tasks without retraining or fine tuning the model. This enables the deep \ac{mpvic} to generalize over multiple tasks where the \ac{rl} approaches are task specific.

Model-free \ac{rl} approaches have been successfully used in \ac{vilc} for robotic manipulation tasks in multiple previous works \cite{martin2019variable, bogdanovic2020learning, varin2019comparison}. Out of which we have chosen the off-policy \ac{rl} algorithm \ac{sac} because of its high sample efficiency. All the three simulation tasks shown in \figref{fig:simulation_tasks} are trained using \ac{sac} implementation from \textit{stable-baselines} \cite{stable-baselines} for \num{500000} time-steps. 

In addition, we compare our approach with the \ac{mbrl} approach \ac{pets} \cite{chua2018deep}. In case of \ac{pets}, the simulation tasks are trained for \num{100000} time-steps. The \ac{pets} policies were trained with same \ac{cem} parameters and cost functions used for the corresponding tasks in our deep \ac{mpvic}. The performance and the transferability of the learned policies in both of these approaches were compared with our \ac{mpvic} approach in \figref{fig:reward_and_transfer}. 

\begin{figure*}[t]
	\centering
	\def\svgwidth{1\linewidth}
	{\fontsize{8}{8}
		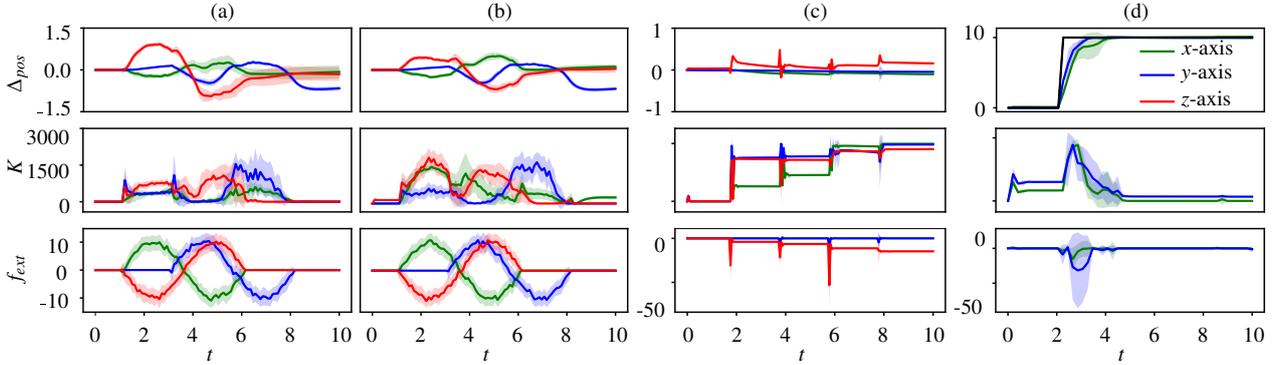}\vspace{-.5cm}
	\caption{Corresponding results from Model-free \ac{rl} policy for the simulation tasks shown in \figref{fig:sims}}
	\label{fig:sims_rl}
\end{figure*}

\begin{figure*}[t]
	\centering
	\def\svgwidth{1\linewidth}
	{\fontsize{8}{8}
		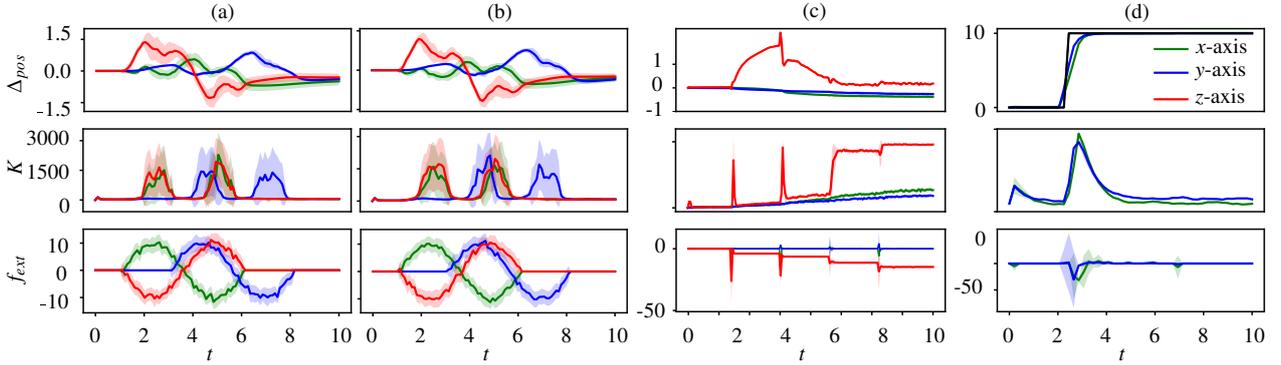}\vspace{-.5cm}
	\caption{Corresponding results from \ac{pets} policy for the simulation tasks shown in \figref{fig:sims}}
	\label{fig:sims_pets}
\end{figure*}

\begin{figure}[t]
    \centering
    \def\svgwidth{1\linewidth}
    {\fontsize{8}{8}
    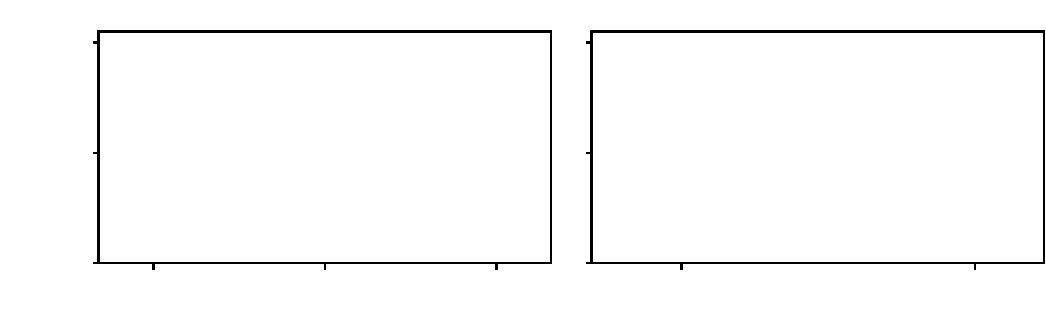}
    \caption{(left) Comparing the normalized value of reward (mean value over 20 trials) obtained using Model-free \ac{rl}, \ac{pets}, and our \ac{mpvic} framework on all the three simulation tasks. (right) Comparing the transferability of the Model-free \ac{rl} and  \ac{pets}  based policy  with our \ac{mpvic} framework based on normalized value of the mean reward over 20 trials. }
    \label{fig:reward_and_transfer}
\end{figure}

\textbf{Performance:} 
The resulting robot behaviour on applying the learned model-free \ac{rl} and \ac{pets} policies on the three simulation tasks are shown in \figref{fig:sims_rl} and \figref{fig:sims_pets} respectively. The reward obtained while applying the learned policies are shown in \figref{fig:reward_and_transfer}. Deep \ac{mpvic} performed better on \textit{task a}, the performance was similar on \textit{task b} and model-free \ac{rl} and \ac{pets} policies performed better on the \textit{task c} by minimizing the stiffness more effectively.

\textbf{Task transferability:}
In order to evaluate how efficiently the policy learned on a task can be transferred to another task, the model-free \ac{rl} and \ac{pets} policies  learned on the simulation \textit{task a} was tested on \textit{task b} and \textit{task c} without retraining the policy/model . The performance of the transferred model-free \ac{rl} and \ac{pets} policies on \textit{task b} and \textit{c} were compared with the corresponding performance of deep \ac{mpvic} using the \ac{penn} model trained on \textit{task a}.  \Figref{fig:reward_and_transfer}-\emph{right} illustrates the transferability of our deep \ac{mpvic} in comparison with \ac{rl}-based approaches, where deep \ac{mpvic} demonstrates the major advantage (green bars). Further, the model-free \ac{rl} and \ac{pets} policies have been retrained to achieve similar performance as our deep \ac{mpvic}. A comparison on the additional \textit{data samples}/\textit{time steps} required for retraining the models/policies for the tasks is shown in Table \ref{tab:computation_time}. The number of additional training samples required is correlated with the computational time. While \ac{rl} approaches demanding additional computational/training time to perform a new task, the proposed deep \ac{mpvic} can be deployed without any additional computational effort.
}
\begin{table}[h!]
	\centering
	\caption{{Comparison on transferability between tasks}}
	\begin{tabular}{m{0.3\linewidth}m{0.3\linewidth}m{0.3\linewidth}m{0.3\linewidth}}
		\cline{1-4}
		 & \multicolumn{3}{c}{\begin{tabular}[c]{@{}c@{}}Training samples ($\times 10^5$)\end{tabular}}\\
		\cline{2-4}
        & \multicolumn{1}{c}{\begin{tabular}[c]{@{}c@{}}\end{tabular}} & \multicolumn{2}{c}{\begin{tabular}[c]{@{}c@{}}Transferability to\end{tabular}} \\
        
		& \multicolumn{1}{c}{\begin{tabular}[c]{@{}c@{}}\textit{Task a}\end{tabular}} & \multicolumn{1}{c}{\begin{tabular}[c]{@{}c@{}}\textit{Task b}\end{tabular}} & \multicolumn{1}{c}{\begin{tabular}[c]{@{}c@{}}\textit{Task c}   \end{tabular} }\\ \hline
		Model-free \ac{rl}	& \multicolumn{1}{c}{$50$}	 & \multicolumn{1}{c}{$38.6$} & \multicolumn{1}{c}{$27.95$} \\ \hline
		\ac{pets} & \multicolumn{1}{c}{$10$}	& \multicolumn{1}{c}{$3.2$} & \multicolumn{1}{c}{$3.9$} \\ \hline
		Our \acs{mpvic} & \multicolumn{1}{c}{$10 $}	& \multicolumn{1}{c}{0} & \multicolumn{1}{c}{0} \\ \hline
	\end{tabular}
	\label{tab:computation_time}
\end{table}

\subsection{Real-World Experiments}
\begin{figure}[t]
    \centering
    \def\svgwidth{1\linewidth}
    {\fontsize{8}{8}
    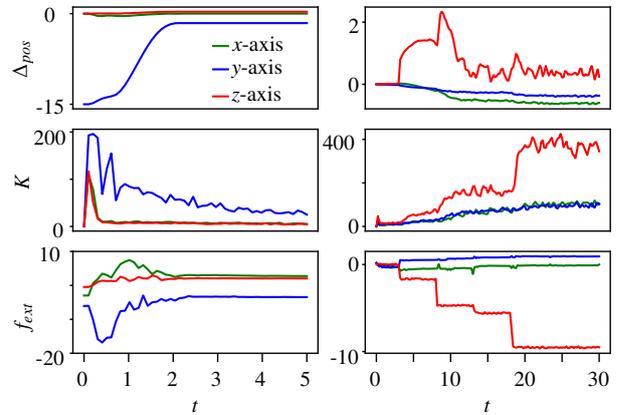}
    \caption{Experiments: (left) task (e), Robot manipulator opening a table drawer. (right) task (d) The robot manipulator with a tray holding its pose while objects are dropped to the tray.}
    \label{fig:exp}
\end{figure}
\textbf{Reacting to falling objects:} The experimental setup is shown in \figref{fig:simulation_tasks} (d) where the robot end-effector is fitted with a tray and four objects of different weights are added to the tray at regular intervals. The optimization objective here is similar to the simulation task (c), the robot is expected to hold a pose while being highly compliant and becoming stiffer with extra weights being introduced to the tray. In \figref{fig:exp} (right-column) the robot with a very low initial stiffness increases the stiffness every instant a new object is introduced to the tray in order to maintain it at the desired pose.

\textbf{Opening a drawer:} The pulling task, which is similar but in the opposite direction of the pushing task. The experimental setup is shown in \figref{fig:simulation_tasks} (e) where the robot is opening a table drawer to a desired position {(\SI{15}{cm} in $x$ direction)} in the Cartesian space. The results shown in \figref{fig:exp} (left-column), shows the impedance adaptation behaviour similar to the pushing task in simulation where the robot increases its stiffness initially to overcome the inertia of the drawer and then decreased once the drawer starts to move closer to the desired position.

\section{Discussion and Limitations}\label{disussions}
\textbf{Discussion:} The deep \ac{mpvic}-based approach presented in the work is evaluated over different tasks in \secref{sec:evaluation} for optimizing impedance adaptation strategies. The objective in all experiments has been consistent in having  high stiffness values for the \ac{vic} only when the task objective demands that. This objective is motivated by human manipulation behavior and can increase the dexterity of the robot while encouraging energy efficient and safe behaviours. In all the evaluation scenarios, both in simulation and experiments, the stiffness adaptation guarantee a high level of compliance unless there is a large deviation from the target position or an external force is applied to it. The modeling approach using \ac{penn} combined with uncertainty targeted exploration has been found to be very useful in learning a generalized unconstrained Cartesian impedance model of the robot. In addition, combining it with \ac{mpc} based optimization has enabled to solve different manipulation tasks demanding stiffness adaptation. The proposed deep \ac{mpvic} approach succeeds in generalising a single model to solve multiple manipulation tasks. The versatility of the  impedance adaptation strategy is evident in the scenarios of impact force from  falling objects, overcoming the inertia of the objects in the pushing and drawer opening tasks respectively. While a majority of robot manipulation tasks rely on trajectory planning and tracking, our approach is not straightforward in solving complex manipulation problems. Nevertheless, it can be combined with a high-level planning approach where the low-level \ac{vic} will modify the given trajectory to ensure compliant behavior. Incorporating such compliant behaviors could improve the manipulation skills, especially in tasks involving contacts. 

{The deep \ac{mpvic} framework was compared with model-free and model-based \ac{rl} approaches utilized successfully in various previous works \cite{martin2019variable, bogdanovic2020learning, varin2019comparison, roveda2020model} to solve complex manipulation tasks. The results show that deep \ac{mpvic} framework is able to achieve similar performance to model-free and model-based \ac{rl} approaches while being highly sample efficient and able to seamlessly transfer the controller between different tasks without any further training of the model. Whereas in model-free and model-based \ac{rl}, transferring policy between different tasks demand relearning the policy on the new task or extensive fine tuning of the existing policy. \ac{pets} shows better task transferability compared to model-free \ac{rl}, this can be justified by the use of a model in \ac{pets} for impedance optimization even though it is not a generalized model as in deep \ac{mpvic}. \ac{rl} has the potential to solve very complex tasks at the expense of high sample complexity. It would be ideal to combine this aspect of \ac{rl} with sample efficiency and easy transferability of the learned controller between tasks as in our deep \ac{mpvic} framework. Further extending the model-based \ac{rl} approaches for \ac{vilc} could be a promising approach in this direction.}

\textbf{Limitations:} Applying our approach to tasks with non-continuous contacts is not possible as the model is not aware of the contact dynamics, which could lead to unstable behavior. {Detecting contact discontinuities and switching to a different contact re-establish policy could be a solution to this issue. Whereas a more general approach could be to learn a model aware of contact constraints, incorporating such constraints into the model state-space is challenging. In the future work we will explore ways to sufficiently incorporate contact constraints to the model to aid faster fine-tuning of the \ac{vilc} for different manipulation tasks.}  In addition, there are limitations inherited from applying \ac{cem} to a real robotic system because of the high computation time, where the trade-off is between optimization performance and the control frequency. {Eventhough \ac{vic} can be operated generally at lower control frequencies, in tasks with complex contact dynamics this might not be sufficient. The level of impedance adaptation or the compliance behaviour can be adjusted by tuning the ${Q}$ and ${R}$ parameters in the cost function \eqref{eq:cost_function}.} However, it is not obvious how to find  optimal values for these parameters. 
  

\section{Conclusion}\label{conclusions}
In this work, we presented a deep \ac{mpvic} approach for compliant manipulation skills for a robotic manipulator by optimizing the impedance parameters. By utilizing \ac{penn}, a Cartesian impedance model of the robot is learned using an exploration strategy maximizing the information gain.  The \ac{penn} dynamic model is coupled with a \ac{cem}-based \ac{mpc} to optimize impedance parameters of a low-level \ac{vic}. We identified an impedance optimization objective-based human manipulation skill and replicated it on a robot manipulator for simplified scenarios in simulations and experiments. 
{The deep \ac{mpvic} was compared with model-free and model-based \ac{rl} approaches in \ac{vilc}.} The approach proved experimentally to be beneficial for solving multiple tasks without any need to relearn the model or policy as opposed to other \ac{vilc} approaches. In the future work, we aim to extend this approach to scenarios with constraints, such as in-contact interaction tasks.





\section*{Acknowledgments}
This work was part of the project ``Dynamic Robot Interaction and Motion Compensation'' funded by the Research Council of Norway under contract number 270941.


\bibliographystyle{model1-num-names}
\bibliography{references}

\end{document}

%% file: acronyms.tex
\usepackage[nolist]{acronym} 

\newacro{vic}[VIC]{Variable Impedance Control}
\newacro{nn}[NN]{Neural Network}
\newacro{penn}[PE]{Probabilistic Ensemble NN}
\newacro{mpc}[MPC]{Model Predictive Control}
\newacro{mpic}[MPIC]{Model Predictive Impedance Control}
\newacro{mpvic}[MPVIC]{Model Predictive Variable Impedance Control}
\newacro{pi2}[PI\textsuperscript{2}]{Policy Improvement with Path Integrals}
\newacro{dmp}[DMP]{Dynamic Movement Primitive}
\newacro{seds}[SEDS]{Stable Estimator of Dynamical Systems}
\newacro{ilc}[ILC]{Iterative  learning  control}
\newacro{gmm}[GMM]{Gaussian Mixture Model}
\newacro{gmr}[GMR]{Gaussian Mixture Regression}
\newacro{gp}[GP]{Gaussian Processes}
\newacro{ppc}[PPC]{Passivity-Preservation Control}
\newacro{tpgmm}[TP-GMM]{Task-Parameterized \ac{gmm}}
\newacro{lfd}[LfD]{Learning from Demonstration}
\newacro{lfhd}[LfD]{Learning-from-human-Demonstration}
\newacro{il}[IL]{Imitation Learning}
\newacro{wls}[WLS]{weighted least-squares}
\newacro{spd}[SPD]{Symmetric Positive Definite}
\newacro{emg}[EMG]{Electromyography}
\newacro{vil}[VIL]{Variable Impedance Learning}
\newacro{vilc}[VILC]{Variable Impedance Learning Control}
\newacro{ai}[AI]{Artificial Intelligent}
\newacro{gadmp}[Ga-DMP]{Geometry-aware Dynamic Movement Primitives}
\newacro{promp}[ProMP]{Probabilistic Movement Primitives}
\newacro{kmp}[KMP]{Kernelized Movement Primitives}
\newacro{dof}[DoF]{Degree of Freedom}
\newacro{msd}[MSD]{virtual-Mass Spring-Damper}
\newacro{gadmp}[GaDMP]{Geometry-aware \ac{dmp}}
\newacro{mbrl}[MBRL]{Model-based Reinforcement Learning}
\newacro{ic}[IC]{Impedance Control}
\newacro{rl}[RL]{Reinforcement Learning}
\newacro{hrc}[HRC]{Human-Robot Collaboration}
\newacro{hri}[HRI]{Human-Robot Interaction}
\newacro{pets}[PETS]{Probabilistic Ensembles with Trajectory Sampling}
\newacro{cem}[CEM]{Cross Entropy Method}
\newacro{sac}[SAC]{Soft Actor Critic}

%% file: macrosNew.tex
\newcommand{\bm}[1]{\boldsymbol{\mathbf{#1}}}

\newcommand{\norm}[1]{\lVert {#1} \rVert}

\newcommand{\figref}[1]{Fig. \ref{#1}}
\newcommand{\figsref}[1]{Figures~\hyperref[#1]{\ref*{#1}}}
\newcommand{\Figref}[1]{Figure~\hyperref[#1]{\ref*{#1}}}

\newcommand{\tabref}[1]{Tab.~\hyperref[#1]{\ref*{#1}}}
\newcommand{\secref}[1]{Section {\ref{#1}}}
\newcommand{\algoref}[1]{Algorithm~\hyperref[#1]{\ref*{#1}}}

\newcommand{\etal}{\MakeLowercase{~\textit{et al.\ }}}

\newlength{\Oldarrayrulewidth}

\definecolor{darkgreen}{rgb}{0.0,0.49,0.19}

\relax
\DeclareFontFamily{U}{eur}{\skewchar\font'177}
\DeclareFontShape{U}{eur}{m}{n}{%
	<-6> eurm5 <6-8> eurm7 <8-> eurm10}{}
\DeclareFontShape{U}{eur}{b}{n}{%
	<-6> eurb5 <6-8> eurb7 <8-> eurb10}{}
\DeclareSymbolFont{ugrf@m}{U}{eur}{m}{n}
\SetSymbolFont{ugrf@m}{bold}{U}{eur}{b}{n}
\DeclareMathSymbol{\upomega}{\mathord}{ugrf@m}{"21}

%% file: figures/framework.pdf_tex
\begingroup%
  \makeatletter%
  \providecommand\color[2][]{%
    \errmessage{(Inkscape) Color is used for the text in Inkscape, but the package 'color.sty' is not loaded}%
    \renewcommand\color[2][]{}%
  }%
  \providecommand\transparent[1]{%
    \errmessage{(Inkscape) Transparency is used (non-zero) for the text in Inkscape, but the package 'transparent.sty' is not loaded}%
    \renewcommand\transparent[1]{}%
  }%
  \providecommand\rotatebox[2]{#2}%
  \newcommand*\fsize{\dimexpr\f@size pt\relax}%
  \newcommand*\lineheight[1]{\fontsize{\fsize}{#1\fsize}\selectfont}%
  \ifx\svgwidth\undefined%
    \setlength{\unitlength}{733.81267301bp}%
    \ifx\svgscale\undefined%
      \relax%
    \else%
      \setlength{\unitlength}{\unitlength * \real{\svgscale}}%
    \fi%
  \else%
    \setlength{\unitlength}{\svgwidth}%
  \fi%
  \global\let\svgwidth\undefined%
  \global\let\svgscale\undefined%
  \makeatother%
  \begin{picture}(1,0.36141038)%
    \lineheight{1}%
    \setlength\tabcolsep{0pt}%
    \put(0,0){\includegraphics[width=\unitlength,page=1]{framework.pdf}}%
    \put(0.16839891,0.3036239){\color[rgb]{0,0,0}\makebox(0,0)[lt]{\lineheight{1.25}\smash{\begin{tabular}[t]{l}$K$\end{tabular}}}}%
    \put(0,0){\includegraphics[width=\unitlength,page=2]{framework.pdf}}%
    \put(0.49328388,0.31243878){\color[rgb]{0,0,0}\makebox(0,0)[lt]{\lineheight{1.25}\smash{\begin{tabular}[t]{l}Robot Inverse\end{tabular}}}}%
    \put(0.5098424,0.29285015){\color[rgb]{0,0,0}\makebox(0,0)[lt]{\lineheight{1.25}\smash{\begin{tabular}[t]{l}Dynamics\end{tabular}}}}%
    \put(0,0){\includegraphics[width=\unitlength,page=3]{framework.pdf}}%
    \put(0.73298455,0.31243878){\color[rgb]{0,0,0}\makebox(0,0)[lt]{\lineheight{1.25}\smash{\begin{tabular}[t]{l}Robot with\end{tabular}}}}%
    \put(0.72602139,0.29285015){\color[rgb]{0,0,0}\makebox(0,0)[lt]{\lineheight{1.25}\smash{\begin{tabular}[t]{l}Environment\end{tabular}}}}%
    \put(0,0){\includegraphics[width=\unitlength,page=4]{framework.pdf}}%
    \put(0.87740484,0.23800196){\color[rgb]{0,0,0}\makebox(0,0)[lt]{\lineheight{1.25}\smash{\begin{tabular}[t]{l}Forward\end{tabular}}}}%
    \put(0.86782477,0.21841332){\color[rgb]{0,0,0}\makebox(0,0)[lt]{\lineheight{1.25}\smash{\begin{tabular}[t]{l}Kinematics\end{tabular}}}}%
    \put(0,0){\includegraphics[width=\unitlength,page=5]{framework.pdf}}%
    \put(0.30834194,0.24485798){\color[rgb]{0,0,0}\makebox(0,0)[lt]{\lineheight{1.25}\smash{\begin{tabular}[t]{l}$D$\end{tabular}}}}%
    \put(0,0){\includegraphics[width=\unitlength,page=6]{framework.pdf}}%
    \put(0.034901,0.09990206){\color[rgb]{0,0,0}\makebox(0,0)[lt]{\lineheight{1.25}\smash{\begin{tabular}[t]{l}Task\end{tabular}}}}%
    \put(0.02008709,0.08031342){\color[rgb]{0,0,0}\makebox(0,0)[lt]{\lineheight{1.25}\smash{\begin{tabular}[t]{l}objective\end{tabular}}}}%
    \put(0,0){\includegraphics[width=\unitlength,page=7]{framework.pdf}}%
    \put(0.55494001,0.09108717){\color[rgb]{0,0,0}\makebox(0,0)[lt]{\lineheight{1.25}\smash{\begin{tabular}[t]{l}$K$\end{tabular}}}}%
    \put(0,0){\includegraphics[width=\unitlength,page=8]{framework.pdf}}%
    \put(0.17801202,0.09271372){\color[rgb]{0,0,0}\makebox(0,0)[lt]{\lineheight{1.25}\smash{\begin{tabular}[t]{l}\ac{cem} + \ac{mpc}\end{tabular}}}}%
    \put(0,0){\includegraphics[width=\unitlength,page=9]{framework.pdf}}%
    \put(0.90089373,0.33202742){\color[rgb]{0,0,0}\makebox(0,0)[lt]{\lineheight{1.25}\smash{\begin{tabular}[t]{l}$f_{ext}$\end{tabular}}}}%
    \put(0.90087843,0.29872674){\color[rgb]{0,0,0}\makebox(0,0)[lt]{\lineheight{1.25}\smash{\begin{tabular}[t]{l}$q, \dot{q}$\end{tabular}}}}%
    \put(0.73145201,0.25661116){\color[rgb]{0,0,0}\makebox(0,0)[lt]{\lineheight{1.25}\smash{\begin{tabular}[t]{l}$\dot{x}$\end{tabular}}}}%
    \put(0.73146732,0.22526934){\color[rgb]{0,0,0}\makebox(0,0)[lt]{\lineheight{1.25}\smash{\begin{tabular}[t]{l}$x$\end{tabular}}}}%
    \put(0.05582762,0.31537708){\color[rgb]{0,0,0}\makebox(0,0)[lt]{\lineheight{1.25}\smash{\begin{tabular}[t]{l}$x^{r}$\end{tabular}}}}%
    \put(0.72967679,0.0215475){\color[rgb]{0,0,0}\makebox(0,0)[lt]{\lineheight{1.25}\smash{\begin{tabular}[t]{l}$x^{r}$\end{tabular}}}}%
    \put(0,0){\includegraphics[width=\unitlength,page=10]{framework.pdf}}%
    \put(0.54081259,0.11263467){\color[rgb]{0,0,0}\makebox(0,0)[lt]{\lineheight{1.25}\smash{\begin{tabular}[t]{l}N\end{tabular}}}}%
    \put(0,0){\includegraphics[width=\unitlength,page=11]{framework.pdf}}%
    \put(0.38057492,0.0920666){\color[rgb]{0,0,0}\makebox(0,0)[lt]{\lineheight{1.25}\smash{\begin{tabular}[t]{l}$s_{t+1}$\end{tabular}}}}%
    \put(0,0){\includegraphics[width=\unitlength,page=12]{framework.pdf}}%
    \put(0.71905607,0.10186092){\color[rgb]{0,0,0}\makebox(0,0)[lt]{\lineheight{1.25}\smash{\begin{tabular}[t]{l}$s_{t}$\end{tabular}}}}%
    \put(0.72014264,0.08716944){\color[rgb]{0,0,0}\makebox(0,0)[lt]{\lineheight{1.25}\smash{\begin{tabular}[t]{l}$u_{t}$\end{tabular}}}}%
    \put(0,0){\includegraphics[width=\unitlength,page=13]{framework.pdf}}%
    \put(0.48008337,0.02359162){\color[rgb]{0,0,0}\makebox(0,0)[lt]{\lineheight{1.25}\smash{\begin{tabular}[t]{l}\acf{penn}\end{tabular}}}}%
    \put(0,0){\includegraphics[width=\unitlength,page=14]{framework.pdf}}%
    \put(0.33735679,0.30558277){\color[rgb]{0,0,0}\makebox(0,0)[lt]{\lineheight{1.25}\smash{\begin{tabular}[t]{l}$M^{-1}$\end{tabular}}}}%
    \put(0,0){\includegraphics[width=\unitlength,page=15]{framework.pdf}}%
    \put(0.72967678,0.18413321){\color[rgb]{0,0,0}\makebox(0,0)[lt]{\lineheight{1.25}\smash{\begin{tabular}[t]{l}$K$\end{tabular}}}}%
    \put(0.24214288,0.33904591){\makebox(0,0)[lt]{\lineheight{1.25}\smash{\begin{tabular}[t]{l}---\end{tabular}}}}%
    \put(0.24407076,0.27454107){\makebox(0,0)[lt]{\lineheight{1.25}\smash{\begin{tabular}[t]{l}---\end{tabular}}}}%
    \put(0.08564041,0.27525974){\makebox(0,0)[lt]{\lineheight{1.25}\smash{\begin{tabular}[t]{l}---\end{tabular}}}}%
  \end{picture}%
\endgroup%

%% file: figures/simulation_tasks.pdf_tex
\begingroup%
  \makeatletter%
  \providecommand\color[2][]{%
    \errmessage{(Inkscape) Color is used for the text in Inkscape, but the package 'color.sty' is not loaded}%
    \renewcommand\color[2][]{}%
  }%
  \providecommand\transparent[1]{%
    \errmessage{(Inkscape) Transparency is used (non-zero) for the text in Inkscape, but the package 'transparent.sty' is not loaded}%
    \renewcommand\transparent[1]{}%
  }%
  \providecommand\rotatebox[2]{#2}%
  \newcommand*\fsize{\dimexpr\f@size pt\relax}%
  \newcommand*\lineheight[1]{\fontsize{\fsize}{#1\fsize}\selectfont}%
  \ifx\svgwidth\undefined%
    \setlength{\unitlength}{517.59323809bp}%
    \ifx\svgscale\undefined%
      \relax%
    \else%
      \setlength{\unitlength}{\unitlength * \real{\svgscale}}%
    \fi%
  \else%
    \setlength{\unitlength}{\svgwidth}%
  \fi%
  \global\let\svgwidth\undefined%
  \global\let\svgscale\undefined%
  \makeatother%
  \begin{picture}(1,0.23803907)%
    \lineheight{1}%
    \setlength\tabcolsep{0pt}%
    \put(0.07188021,0.22816497){\makebox(0,0)[lt]{\lineheight{1.25}\smash{\begin{tabular}[t]{l}(a)\end{tabular}}}}%
    \put(0.24793501,0.22754358){\makebox(0,0)[lt]{\lineheight{1.25}\smash{\begin{tabular}[t]{l}(b)\end{tabular}}}}%
    \put(0.4315724,0.22864397){\makebox(0,0)[lt]{\lineheight{1.25}\smash{\begin{tabular}[t]{l}(c)\end{tabular}}}}%
    \put(0.61555634,0.22771213){\makebox(0,0)[lt]{\lineheight{1.25}\smash{\begin{tabular}[t]{l}(d)\end{tabular}}}}%
    \put(0.85376605,0.22723024){\makebox(0,0)[lt]{\lineheight{1.25}\smash{\begin{tabular}[t]{l}(e)\end{tabular}}}}%
    \put(0,0){\includegraphics[width=\unitlength,page=1]{simulation_tasks.pdf}}%
    \put(0.46459535,0.01255109){\color[rgb]{0,0,0}\makebox(0,0)[lt]{\lineheight{1.25}\smash{\begin{tabular}[t]{l}Goal\end{tabular}}}}%
    \put(0,0){\includegraphics[width=\unitlength,page=2]{simulation_tasks.pdf}}%
    \put(0.02228466,0.09359176){\makebox(0,0)[lt]{\lineheight{1.25}\smash{\begin{tabular}[t]{l}\textcolor{white}{Force}\end{tabular}}}}%
    \put(0,0){\includegraphics[width=\unitlength,page=3]{simulation_tasks.pdf}}%
    \put(0.1194123,0.10992818){\makebox(0,0)[lt]{\lineheight{1.25}\smash{\begin{tabular}[t]{l}\textcolor{white}{Goal}\end{tabular}}}}%
    \put(0,0){\includegraphics[width=\unitlength,page=4]{simulation_tasks.pdf}}%
  \end{picture}%
\endgroup%

%% file: figures/sims_new.pdf_tex
\begingroup%
  \makeatletter%
  \providecommand\color[2][]{%
    \errmessage{(Inkscape) Color is used for the text in Inkscape, but the package 'color.sty' is not loaded}%
    \renewcommand\color[2][]{}%
  }%
  \providecommand\transparent[1]{%
    \errmessage{(Inkscape) Transparency is used (non-zero) for the text in Inkscape, but the package 'transparent.sty' is not loaded}%
    \renewcommand\transparent[1]{}%
  }%
  \providecommand\rotatebox[2]{#2}%
  \newcommand*\fsize{\dimexpr\f@size pt\relax}%
  \newcommand*\lineheight[1]{\fontsize{\fsize}{#1\fsize}\selectfont}%
  \ifx\svgwidth\undefined%
    \setlength{\unitlength}{646.53567903bp}%
    \ifx\svgscale\undefined%
      \relax%
    \else%
      \setlength{\unitlength}{\unitlength * \real{\svgscale}}%
    \fi%
  \else%
    \setlength{\unitlength}{\svgwidth}%
  \fi%
  \global\let\svgwidth\undefined%
  \global\let\svgscale\undefined%
  \makeatother%
  \begin{picture}(1,0.28369769)%
    \lineheight{1}%
    \setlength\tabcolsep{0pt}%
    \put(0,0){\includegraphics[width=\unitlength,page=1]{sims_new.pdf}}%
    \put(0.02359658,0.19843402){\color[rgb]{0,0,0}\makebox(0,0)[lt]{\lineheight{1.25}\smash{\begin{tabular}[t]{l}1.5\end{tabular}}}}%
    \put(0,0){\includegraphics[width=\unitlength,page=2]{sims_new.pdf}}%
    \put(0.02343535,0.2289156){\color[rgb]{0,0,0}\makebox(0,0)[lt]{\lineheight{1.25}\smash{\begin{tabular}[t]{l}0.0\end{tabular}}}}%
    \put(0,0){\includegraphics[width=\unitlength,page=3]{sims_new.pdf}}%
    \put(0.02359658,0.25939719){\color[rgb]{0,0,0}\makebox(0,0)[lt]{\lineheight{1.25}\smash{\begin{tabular}[t]{l}1.5\end{tabular}}}}%
    \put(0.15668174,0.27742444){\color[rgb]{0,0,0}\makebox(0,0)[lt]{\lineheight{1.25}\smash{\begin{tabular}[t]{l}(a)\end{tabular}}}}%
    \put(0.37780743,0.27742444){\color[rgb]{0,0,0}\makebox(0,0)[lt]{\lineheight{1.25}\smash{\begin{tabular}[t]{l}(b)\end{tabular}}}}%
    \put(0.63098243,0.27742444){\color[rgb]{0,0,0}\makebox(0,0)[lt]{\lineheight{1.25}\smash{\begin{tabular}[t]{l}(c)\end{tabular}}}}%
    \put(0.88694873,0.27742444){\color[rgb]{0,0,0}\makebox(0,0)[lt]{\lineheight{1.25}\smash{\begin{tabular}[t]{l}(d)\end{tabular}}}}%
    \put(0.00508436,0.21687144){\color[rgb]{0,0,0}\rotatebox{90}{\makebox(0,0)[lt]{\lineheight{1.25}\smash{\begin{tabular}[t]{l}$\Delta_{pos}$\end{tabular}}}}}%
    \put(0,0){\includegraphics[width=\unitlength,page=4]{sims_new.pdf}}%
    \put(0.03540129,0.12025707){\color[rgb]{0,0,0}\makebox(0,0)[lt]{\lineheight{1.25}\smash{\begin{tabular}[t]{l}0\end{tabular}}}}%
    \put(0,0){\includegraphics[width=\unitlength,page=5]{sims_new.pdf}}%
    \put(0.0091508,0.14780172){\color[rgb]{0,0,0}\makebox(0,0)[lt]{\lineheight{1.25}\smash{\begin{tabular}[t]{l}1500\end{tabular}}}}%
    \put(0,0){\includegraphics[width=\unitlength,page=6]{sims_new.pdf}}%
    \put(0.0091508,0.17534638){\color[rgb]{0,0,0}\makebox(0,0)[lt]{\lineheight{1.25}\smash{\begin{tabular}[t]{l}3000\end{tabular}}}}%
    \put(0,0){\includegraphics[width=\unitlength,page=7]{sims_new.pdf}}%
    \put(0.05988427,0.01620028){\color[rgb]{0,0,0}\makebox(0,0)[lt]{\lineheight{1.25}\smash{\begin{tabular}[t]{l}0\end{tabular}}}}%
    \put(0,0){\includegraphics[width=\unitlength,page=8]{sims_new.pdf}}%
    \put(0.09883804,0.01620028){\color[rgb]{0,0,0}\makebox(0,0)[lt]{\lineheight{1.25}\smash{\begin{tabular}[t]{l}2\end{tabular}}}}%
    \put(0,0){\includegraphics[width=\unitlength,page=9]{sims_new.pdf}}%
    \put(0.13779183,0.01620028){\color[rgb]{0,0,0}\makebox(0,0)[lt]{\lineheight{1.25}\smash{\begin{tabular}[t]{l}4\end{tabular}}}}%
    \put(0,0){\includegraphics[width=\unitlength,page=10]{sims_new.pdf}}%
    \put(0.1767456,0.01620028){\color[rgb]{0,0,0}\makebox(0,0)[lt]{\lineheight{1.25}\smash{\begin{tabular}[t]{l}6\end{tabular}}}}%
    \put(0,0){\includegraphics[width=\unitlength,page=11]{sims_new.pdf}}%
    \put(0.21569937,0.01620028){\color[rgb]{0,0,0}\makebox(0,0)[lt]{\lineheight{1.25}\smash{\begin{tabular}[t]{l}8\end{tabular}}}}%
    \put(0,0){\includegraphics[width=\unitlength,page=12]{sims_new.pdf}}%
    \put(0.25221146,0.01620028){\color[rgb]{0,0,0}\makebox(0,0)[lt]{\lineheight{1.25}\smash{\begin{tabular}[t]{l}10\end{tabular}}}}%
    \put(0,0){\includegraphics[width=\unitlength,page=13]{sims_new.pdf}}%
    \put(0.02587729,0.04348314){\color[rgb]{0,0,0}\makebox(0,0)[lt]{\lineheight{1.25}\smash{\begin{tabular}[t]{l}10\end{tabular}}}}%
    \put(0,0){\includegraphics[width=\unitlength,page=14]{sims_new.pdf}}%
    \put(0.03308123,0.06578872){\color[rgb]{0,0,0}\makebox(0,0)[lt]{\lineheight{1.25}\smash{\begin{tabular}[t]{l}0\end{tabular}}}}%
    \put(0,0){\includegraphics[width=\unitlength,page=15]{sims_new.pdf}}%
    \put(0.02587729,0.08809431){\color[rgb]{0,0,0}\makebox(0,0)[lt]{\lineheight{1.25}\smash{\begin{tabular}[t]{l}10\end{tabular}}}}%
    \put(0,0){\includegraphics[width=\unitlength,page=16]{sims_new.pdf}}%
    \put(0.28115535,0.01615563){\color[rgb]{0,0,0}\makebox(0,0)[lt]{\lineheight{1.25}\smash{\begin{tabular}[t]{l}0\end{tabular}}}}%
    \put(0,0){\includegraphics[width=\unitlength,page=17]{sims_new.pdf}}%
    \put(0.32005238,0.01615563){\color[rgb]{0,0,0}\makebox(0,0)[lt]{\lineheight{1.25}\smash{\begin{tabular}[t]{l}2\end{tabular}}}}%
    \put(0,0){\includegraphics[width=\unitlength,page=18]{sims_new.pdf}}%
    \put(0.35894945,0.01615563){\color[rgb]{0,0,0}\makebox(0,0)[lt]{\lineheight{1.25}\smash{\begin{tabular}[t]{l}4\end{tabular}}}}%
    \put(0,0){\includegraphics[width=\unitlength,page=19]{sims_new.pdf}}%
    \put(0.39784648,0.01615563){\color[rgb]{0,0,0}\makebox(0,0)[lt]{\lineheight{1.25}\smash{\begin{tabular}[t]{l}6\end{tabular}}}}%
    \put(0,0){\includegraphics[width=\unitlength,page=20]{sims_new.pdf}}%
    \put(0.43674354,0.01615563){\color[rgb]{0,0,0}\makebox(0,0)[lt]{\lineheight{1.25}\smash{\begin{tabular}[t]{l}8\end{tabular}}}}%
    \put(0,0){\includegraphics[width=\unitlength,page=21]{sims_new.pdf}}%
    \put(0.47320238,0.01615563){\color[rgb]{0,0,0}\makebox(0,0)[lt]{\lineheight{1.25}\smash{\begin{tabular}[t]{l}10\end{tabular}}}}%
    \put(0.4971544,0.03616871){\color[rgb]{0,0,0}\makebox(0,0)[lt]{\lineheight{1.25}\smash{\begin{tabular}[t]{l}-50\end{tabular}}}}%
    \put(0,0){\includegraphics[width=\unitlength,page=22]{sims_new.pdf}}%
    \put(0.00538171,0.15056314){\color[rgb]{0,0,0}\rotatebox{90}{\makebox(0,0)[lt]{\lineheight{1.25}\smash{\begin{tabular}[t]{l}$K$\end{tabular}}}}}%
    \put(0.00508436,0.06254598){\color[rgb]{0,0,0}\rotatebox{90}{\makebox(0,0)[lt]{\lineheight{1.25}\smash{\begin{tabular}[t]{l}$f_{ext}$\end{tabular}}}}}%
    \put(0.51176003,0.08847607){\color[rgb]{0,0,0}\makebox(0,0)[lt]{\lineheight{1.25}\smash{\begin{tabular}[t]{l}0\end{tabular}}}}%
    \put(0.50936889,0.22727065){\color[rgb]{0,0,0}\makebox(0,0)[lt]{\lineheight{1.25}\smash{\begin{tabular}[t]{l}0\end{tabular}}}}%
    \put(0.50956732,0.26112299){\color[rgb]{0,0,0}\makebox(0,0)[lt]{\lineheight{1.25}\smash{\begin{tabular}[t]{l}1\end{tabular}}}}%
    \put(0.50448029,0.19839611){\color[rgb]{0,0,0}\makebox(0,0)[lt]{\lineheight{1.25}\smash{\begin{tabular}[t]{l}-1\end{tabular}}}}%
    \put(0.75105374,0.03889419){\color[rgb]{0,0,0}\makebox(0,0)[lt]{\lineheight{1.25}\smash{\begin{tabular}[t]{l}-50\end{tabular}}}}%
    \put(0.76794389,0.09350403){\color[rgb]{0,0,0}\makebox(0,0)[lt]{\lineheight{1.25}\smash{\begin{tabular}[t]{l}0\end{tabular}}}}%
    \put(0.76794839,0.19770206){\color[rgb]{0,0,0}\makebox(0,0)[lt]{\lineheight{1.25}\smash{\begin{tabular}[t]{l}0\end{tabular}}}}%
    \put(0.7584176,0.2600482){\color[rgb]{0,0,0}\makebox(0,0)[lt]{\lineheight{1.25}\smash{\begin{tabular}[t]{l}10\end{tabular}}}}%
    \put(0.37445434,0.00121361){\color[rgb]{0,0,0}\makebox(0,0)[lt]{\lineheight{1.25}\smash{\begin{tabular}[t]{l}$t$\end{tabular}}}}%
    \put(0.15332864,0.00121361){\color[rgb]{0,0,0}\makebox(0,0)[lt]{\lineheight{1.25}\smash{\begin{tabular}[t]{l}$t$\end{tabular}}}}%
    \put(0.53438444,0.01615582){\color[rgb]{0,0,0}\makebox(0,0)[lt]{\lineheight{1.25}\smash{\begin{tabular}[t]{l}0\end{tabular}}}}%
    \put(0.57333821,0.01609736){\color[rgb]{0,0,0}\makebox(0,0)[lt]{\lineheight{1.25}\smash{\begin{tabular}[t]{l}2\end{tabular}}}}%
    \put(0.61229199,0.01615179){\color[rgb]{0,0,0}\makebox(0,0)[lt]{\lineheight{1.25}\smash{\begin{tabular}[t]{l}4\end{tabular}}}}%
    \put(0.65124576,0.01615582){\color[rgb]{0,0,0}\makebox(0,0)[lt]{\lineheight{1.25}\smash{\begin{tabular}[t]{l}6\end{tabular}}}}%
    \put(0.69019953,0.01615582){\color[rgb]{0,0,0}\makebox(0,0)[lt]{\lineheight{1.25}\smash{\begin{tabular}[t]{l}8\end{tabular}}}}%
    \put(0.72671161,0.01615582){\color[rgb]{0,0,0}\makebox(0,0)[lt]{\lineheight{1.25}\smash{\begin{tabular}[t]{l}10\end{tabular}}}}%
    \put(0.62684709,0.00121361){\color[rgb]{0,0,0}\makebox(0,0)[lt]{\lineheight{1.25}\smash{\begin{tabular}[t]{l}$t$\end{tabular}}}}%
    \put(0.79023469,0.01615582){\color[rgb]{0,0,0}\makebox(0,0)[lt]{\lineheight{1.25}\smash{\begin{tabular}[t]{l}0\end{tabular}}}}%
    \put(0.82918846,0.01609736){\color[rgb]{0,0,0}\makebox(0,0)[lt]{\lineheight{1.25}\smash{\begin{tabular}[t]{l}2\end{tabular}}}}%
    \put(0.86814224,0.01615179){\color[rgb]{0,0,0}\makebox(0,0)[lt]{\lineheight{1.25}\smash{\begin{tabular}[t]{l}4\end{tabular}}}}%
    \put(0.90709601,0.01615582){\color[rgb]{0,0,0}\makebox(0,0)[lt]{\lineheight{1.25}\smash{\begin{tabular}[t]{l}6\end{tabular}}}}%
    \put(0.94604978,0.01615582){\color[rgb]{0,0,0}\makebox(0,0)[lt]{\lineheight{1.25}\smash{\begin{tabular}[t]{l}8\end{tabular}}}}%
    \put(0.98256186,0.01615582){\color[rgb]{0,0,0}\makebox(0,0)[lt]{\lineheight{1.25}\smash{\begin{tabular}[t]{l}10\end{tabular}}}}%
    \put(0.88269734,0.00121361){\color[rgb]{0,0,0}\makebox(0,0)[lt]{\lineheight{1.25}\smash{\begin{tabular}[t]{l}$t$\end{tabular}}}}%
    \put(0,0){\includegraphics[width=\unitlength,page=23]{sims_new.pdf}}%
    \put(0.93302025,0.23801625){\color[rgb]{0,0,0}\makebox(0,0)[lt]{\lineheight{1.25}\smash{\begin{tabular}[t]{l}$x$-axis\end{tabular}}}}%
    \put(0,0){\includegraphics[width=\unitlength,page=24]{sims_new.pdf}}%
    \put(0.93302025,0.22358188){\color[rgb]{0,0,0}\makebox(0,0)[lt]{\lineheight{1.25}\smash{\begin{tabular}[t]{l}$y$-axis\end{tabular}}}}%
    \put(0,0){\includegraphics[width=\unitlength,page=25]{sims_new.pdf}}%
    \put(0.93302025,0.20682746){\color[rgb]{0,0,0}\makebox(0,0)[lt]{\lineheight{1.25}\smash{\begin{tabular}[t]{l}$z$-axis\end{tabular}}}}%
  \end{picture}%
\endgroup%

%% file: figures/sims_RL.pdf_tex
\begingroup%
  \makeatletter%
  \providecommand\color[2][]{%
    \errmessage{(Inkscape) Color is used for the text in Inkscape, but the package 'color.sty' is not loaded}%
    \renewcommand\color[2][]{}%
  }%
  \providecommand\transparent[1]{%
    \errmessage{(Inkscape) Transparency is used (non-zero) for the text in Inkscape, but the package 'transparent.sty' is not loaded}%
    \renewcommand\transparent[1]{}%
  }%
  \providecommand\rotatebox[2]{#2}%
  \newcommand*\fsize{\dimexpr\f@size pt\relax}%
  \newcommand*\lineheight[1]{\fontsize{\fsize}{#1\fsize}\selectfont}%
  \ifx\svgwidth\undefined%
    \setlength{\unitlength}{646.53564189bp}%
    \ifx\svgscale\undefined%
      \relax%
    \else%
      \setlength{\unitlength}{\unitlength * \real{\svgscale}}%
    \fi%
  \else%
    \setlength{\unitlength}{\svgwidth}%
  \fi%
  \global\let\svgwidth\undefined%
  \global\let\svgscale\undefined%
  \makeatother%
  \begin{picture}(1,0.2836977)%
    \lineheight{1}%
    \setlength\tabcolsep{0pt}%
    \put(0.02359658,0.19843401){\color[rgb]{0,0,0}\makebox(0,0)[lt]{\lineheight{1.25}\smash{\begin{tabular}[t]{l}1.5\end{tabular}}}}%
    \put(0,0){\includegraphics[width=\unitlength,page=1]{sims_RL.pdf}}%
    \put(0.02343535,0.22891562){\color[rgb]{0,0,0}\makebox(0,0)[lt]{\lineheight{1.25}\smash{\begin{tabular}[t]{l}0.0\end{tabular}}}}%
    \put(0.02359658,0.25939719){\color[rgb]{0,0,0}\makebox(0,0)[lt]{\lineheight{1.25}\smash{\begin{tabular}[t]{l}1.5\end{tabular}}}}%
    \put(0.15668175,0.27742446){\color[rgb]{0,0,0}\makebox(0,0)[lt]{\lineheight{1.25}\smash{\begin{tabular}[t]{l}(a)\end{tabular}}}}%
    \put(0.37780745,0.27742446){\color[rgb]{0,0,0}\makebox(0,0)[lt]{\lineheight{1.25}\smash{\begin{tabular}[t]{l}(b)\end{tabular}}}}%
    \put(0.63098245,0.27742446){\color[rgb]{0,0,0}\makebox(0,0)[lt]{\lineheight{1.25}\smash{\begin{tabular}[t]{l}(c)\end{tabular}}}}%
    \put(0.88694877,0.27742446){\color[rgb]{0,0,0}\makebox(0,0)[lt]{\lineheight{1.25}\smash{\begin{tabular}[t]{l}(d)\end{tabular}}}}%
    \put(0.00508436,0.21687147){\color[rgb]{0,0,0}\rotatebox{90}{\makebox(0,0)[lt]{\lineheight{1.25}\smash{\begin{tabular}[t]{l}$\Delta_{pos}$\end{tabular}}}}}%
    \put(0.03540129,0.12025708){\color[rgb]{0,0,0}\makebox(0,0)[lt]{\lineheight{1.25}\smash{\begin{tabular}[t]{l}0\end{tabular}}}}%
    \put(0.0091508,0.14780174){\color[rgb]{0,0,0}\makebox(0,0)[lt]{\lineheight{1.25}\smash{\begin{tabular}[t]{l}1500\end{tabular}}}}%
    \put(0.0091508,0.1753464){\color[rgb]{0,0,0}\makebox(0,0)[lt]{\lineheight{1.25}\smash{\begin{tabular}[t]{l}3000\end{tabular}}}}%
    \put(0.05988428,0.01620028){\color[rgb]{0,0,0}\makebox(0,0)[lt]{\lineheight{1.25}\smash{\begin{tabular}[t]{l}0\end{tabular}}}}%
    \put(0.09883805,0.01620028){\color[rgb]{0,0,0}\makebox(0,0)[lt]{\lineheight{1.25}\smash{\begin{tabular}[t]{l}2\end{tabular}}}}%
    \put(0.13779184,0.01620028){\color[rgb]{0,0,0}\makebox(0,0)[lt]{\lineheight{1.25}\smash{\begin{tabular}[t]{l}4\end{tabular}}}}%
    \put(0.17674561,0.01620028){\color[rgb]{0,0,0}\makebox(0,0)[lt]{\lineheight{1.25}\smash{\begin{tabular}[t]{l}6\end{tabular}}}}%
    \put(0.21569938,0.01620028){\color[rgb]{0,0,0}\makebox(0,0)[lt]{\lineheight{1.25}\smash{\begin{tabular}[t]{l}8\end{tabular}}}}%
    \put(0.25221148,0.01620028){\color[rgb]{0,0,0}\makebox(0,0)[lt]{\lineheight{1.25}\smash{\begin{tabular}[t]{l}10\end{tabular}}}}%
    \put(0.0258773,0.04348315){\color[rgb]{0,0,0}\makebox(0,0)[lt]{\lineheight{1.25}\smash{\begin{tabular}[t]{l}10\end{tabular}}}}%
    \put(0,0){\includegraphics[width=\unitlength,page=2]{sims_RL.pdf}}%
    \put(0.03308124,0.06578872){\color[rgb]{0,0,0}\makebox(0,0)[lt]{\lineheight{1.25}\smash{\begin{tabular}[t]{l}0\end{tabular}}}}%
    \put(0.0258773,0.08809433){\color[rgb]{0,0,0}\makebox(0,0)[lt]{\lineheight{1.25}\smash{\begin{tabular}[t]{l}10\end{tabular}}}}%
    \put(0.28115536,0.0161556){\color[rgb]{0,0,0}\makebox(0,0)[lt]{\lineheight{1.25}\smash{\begin{tabular}[t]{l}0\end{tabular}}}}%
    \put(0.3200524,0.0161556){\color[rgb]{0,0,0}\makebox(0,0)[lt]{\lineheight{1.25}\smash{\begin{tabular}[t]{l}2\end{tabular}}}}%
    \put(0.35894947,0.0161556){\color[rgb]{0,0,0}\makebox(0,0)[lt]{\lineheight{1.25}\smash{\begin{tabular}[t]{l}4\end{tabular}}}}%
    \put(0.3978465,0.0161556){\color[rgb]{0,0,0}\makebox(0,0)[lt]{\lineheight{1.25}\smash{\begin{tabular}[t]{l}6\end{tabular}}}}%
    \put(0.43674356,0.0161556){\color[rgb]{0,0,0}\makebox(0,0)[lt]{\lineheight{1.25}\smash{\begin{tabular}[t]{l}8\end{tabular}}}}%
    \put(0.47320241,0.0161556){\color[rgb]{0,0,0}\makebox(0,0)[lt]{\lineheight{1.25}\smash{\begin{tabular}[t]{l}10\end{tabular}}}}%
    \put(0.49715441,0.03616872){\color[rgb]{0,0,0}\makebox(0,0)[lt]{\lineheight{1.25}\smash{\begin{tabular}[t]{l}-50\end{tabular}}}}%
    \put(0.00538171,0.15056317){\color[rgb]{0,0,0}\rotatebox{90}{\makebox(0,0)[lt]{\lineheight{1.25}\smash{\begin{tabular}[t]{l}$K$\end{tabular}}}}}%
    \put(0.00508436,0.06254599){\color[rgb]{0,0,0}\rotatebox{90}{\makebox(0,0)[lt]{\lineheight{1.25}\smash{\begin{tabular}[t]{l}$f_{ext}$\end{tabular}}}}}%
    \put(0.51176005,0.08847607){\color[rgb]{0,0,0}\makebox(0,0)[lt]{\lineheight{1.25}\smash{\begin{tabular}[t]{l}0\end{tabular}}}}%
    \put(0.50936891,0.22727065){\color[rgb]{0,0,0}\makebox(0,0)[lt]{\lineheight{1.25}\smash{\begin{tabular}[t]{l}0\end{tabular}}}}%
    \put(0.50956733,0.26112298){\color[rgb]{0,0,0}\makebox(0,0)[lt]{\lineheight{1.25}\smash{\begin{tabular}[t]{l}1\end{tabular}}}}%
    \put(0.50448032,0.1983961){\color[rgb]{0,0,0}\makebox(0,0)[lt]{\lineheight{1.25}\smash{\begin{tabular}[t]{l}-1\end{tabular}}}}%
    \put(0.75105377,0.03889417){\color[rgb]{0,0,0}\makebox(0,0)[lt]{\lineheight{1.25}\smash{\begin{tabular}[t]{l}-50\end{tabular}}}}%
    \put(0.76794392,0.09350404){\color[rgb]{0,0,0}\makebox(0,0)[lt]{\lineheight{1.25}\smash{\begin{tabular}[t]{l}0\end{tabular}}}}%
    \put(0.76794842,0.19770205){\color[rgb]{0,0,0}\makebox(0,0)[lt]{\lineheight{1.25}\smash{\begin{tabular}[t]{l}0\end{tabular}}}}%
    \put(0.75841763,0.26004819){\color[rgb]{0,0,0}\makebox(0,0)[lt]{\lineheight{1.25}\smash{\begin{tabular}[t]{l}10\end{tabular}}}}%
    \put(0.37445436,0.00121361){\color[rgb]{0,0,0}\makebox(0,0)[lt]{\lineheight{1.25}\smash{\begin{tabular}[t]{l}$t$\end{tabular}}}}%
    \put(0.15332865,0.00121361){\color[rgb]{0,0,0}\makebox(0,0)[lt]{\lineheight{1.25}\smash{\begin{tabular}[t]{l}$t$\end{tabular}}}}%
    \put(0.53438445,0.01615582){\color[rgb]{0,0,0}\makebox(0,0)[lt]{\lineheight{1.25}\smash{\begin{tabular}[t]{l}0\end{tabular}}}}%
    \put(0.57333823,0.01609738){\color[rgb]{0,0,0}\makebox(0,0)[lt]{\lineheight{1.25}\smash{\begin{tabular}[t]{l}2\end{tabular}}}}%
    \put(0.61229201,0.01615179){\color[rgb]{0,0,0}\makebox(0,0)[lt]{\lineheight{1.25}\smash{\begin{tabular}[t]{l}4\end{tabular}}}}%
    \put(0.65124578,0.01615582){\color[rgb]{0,0,0}\makebox(0,0)[lt]{\lineheight{1.25}\smash{\begin{tabular}[t]{l}6\end{tabular}}}}%
    \put(0.69019956,0.01615582){\color[rgb]{0,0,0}\makebox(0,0)[lt]{\lineheight{1.25}\smash{\begin{tabular}[t]{l}8\end{tabular}}}}%
    \put(0.72671164,0.01615582){\color[rgb]{0,0,0}\makebox(0,0)[lt]{\lineheight{1.25}\smash{\begin{tabular}[t]{l}10\end{tabular}}}}%
    \put(0.62684712,0.00121361){\color[rgb]{0,0,0}\makebox(0,0)[lt]{\lineheight{1.25}\smash{\begin{tabular}[t]{l}$t$\end{tabular}}}}%
    \put(0.79023472,0.01615582){\color[rgb]{0,0,0}\makebox(0,0)[lt]{\lineheight{1.25}\smash{\begin{tabular}[t]{l}0\end{tabular}}}}%
    \put(0.8291885,0.01609738){\color[rgb]{0,0,0}\makebox(0,0)[lt]{\lineheight{1.25}\smash{\begin{tabular}[t]{l}2\end{tabular}}}}%
    \put(0.86814227,0.01615179){\color[rgb]{0,0,0}\makebox(0,0)[lt]{\lineheight{1.25}\smash{\begin{tabular}[t]{l}4\end{tabular}}}}%
    \put(0.90709605,0.01615582){\color[rgb]{0,0,0}\makebox(0,0)[lt]{\lineheight{1.25}\smash{\begin{tabular}[t]{l}6\end{tabular}}}}%
    \put(0.94604982,0.01615582){\color[rgb]{0,0,0}\makebox(0,0)[lt]{\lineheight{1.25}\smash{\begin{tabular}[t]{l}8\end{tabular}}}}%
    \put(0.98256191,0.01615582){\color[rgb]{0,0,0}\makebox(0,0)[lt]{\lineheight{1.25}\smash{\begin{tabular}[t]{l}10\end{tabular}}}}%
    \put(0.88269738,0.00121361){\color[rgb]{0,0,0}\makebox(0,0)[lt]{\lineheight{1.25}\smash{\begin{tabular}[t]{l}$t$\end{tabular}}}}%
    \put(0,0){\includegraphics[width=\unitlength,page=3]{sims_RL.pdf}}%
    \put(0.93302029,0.24729653){\color[rgb]{0,0,0}\makebox(0,0)[lt]{\lineheight{1.25}\smash{\begin{tabular}[t]{l}$x$-axis\end{tabular}}}}%
    \put(0,0){\includegraphics[width=\unitlength,page=4]{sims_RL.pdf}}%
    \put(0.93302029,0.22822202){\color[rgb]{0,0,0}\makebox(0,0)[lt]{\lineheight{1.25}\smash{\begin{tabular}[t]{l}$y$-axis\end{tabular}}}}%
    \put(0,0){\includegraphics[width=\unitlength,page=5]{sims_RL.pdf}}%
    \put(0.93302029,0.20682748){\color[rgb]{0,0,0}\makebox(0,0)[lt]{\lineheight{1.25}\smash{\begin{tabular}[t]{l}$z$-axis\end{tabular}}}}%
    \put(0,0){\includegraphics[width=\unitlength,page=6]{sims_RL.pdf}}%
  \end{picture}%
\endgroup%

%% file: figures/sims_pets.pdf_tex
\begingroup%
  \makeatletter%
  \providecommand\color[2][]{%
    \errmessage{(Inkscape) Color is used for the text in Inkscape, but the package 'color.sty' is not loaded}%
    \renewcommand\color[2][]{}%
  }%
  \providecommand\transparent[1]{%
    \errmessage{(Inkscape) Transparency is used (non-zero) for the text in Inkscape, but the package 'transparent.sty' is not loaded}%
    \renewcommand\transparent[1]{}%
  }%
  \providecommand\rotatebox[2]{#2}%
  \newcommand*\fsize{\dimexpr\f@size pt\relax}%
  \newcommand*\lineheight[1]{\fontsize{\fsize}{#1\fsize}\selectfont}%
  \ifx\svgwidth\undefined%
    \setlength{\unitlength}{646.33455213bp}%
    \ifx\svgscale\undefined%
      \relax%
    \else%
      \setlength{\unitlength}{\unitlength * \real{\svgscale}}%
    \fi%
  \else%
    \setlength{\unitlength}{\svgwidth}%
  \fi%
  \global\let\svgwidth\undefined%
  \global\let\svgscale\undefined%
  \makeatother%
  \begin{picture}(1,0.28402828)%
    \lineheight{1}%
    \setlength\tabcolsep{0pt}%
    \put(0.02360393,0.19849575){\color[rgb]{0,0,0}\makebox(0,0)[lt]{\lineheight{1.25}\smash{\begin{tabular}[t]{l}1.5\end{tabular}}}}%
    \put(0,0){\includegraphics[width=\unitlength,page=1]{sims_pets.pdf}}%
    \put(0.02344265,0.22898684){\color[rgb]{0,0,0}\makebox(0,0)[lt]{\lineheight{1.25}\smash{\begin{tabular}[t]{l}0.0\end{tabular}}}}%
    \put(0.02360393,0.2594779){\color[rgb]{0,0,0}\makebox(0,0)[lt]{\lineheight{1.25}\smash{\begin{tabular}[t]{l}1.5\end{tabular}}}}%
    \put(0.1567305,0.27751077){\color[rgb]{0,0,0}\makebox(0,0)[lt]{\lineheight{1.25}\smash{\begin{tabular}[t]{l}(a)\end{tabular}}}}%
    \put(0.37792499,0.27751077){\color[rgb]{0,0,0}\makebox(0,0)[lt]{\lineheight{1.25}\smash{\begin{tabular}[t]{l}(b)\end{tabular}}}}%
    \put(0.00508594,0.21693894){\color[rgb]{0,0,0}\rotatebox{90}{\makebox(0,0)[lt]{\lineheight{1.25}\smash{\begin{tabular}[t]{l}$\Delta_{pos}$\end{tabular}}}}}%
    \put(0.03541231,0.1202945){\color[rgb]{0,0,0}\makebox(0,0)[lt]{\lineheight{1.25}\smash{\begin{tabular}[t]{l}0\end{tabular}}}}%
    \put(0.00915364,0.14784773){\color[rgb]{0,0,0}\makebox(0,0)[lt]{\lineheight{1.25}\smash{\begin{tabular}[t]{l}1500\end{tabular}}}}%
    \put(0.00915364,0.17540096){\color[rgb]{0,0,0}\makebox(0,0)[lt]{\lineheight{1.25}\smash{\begin{tabular}[t]{l}3000\end{tabular}}}}%
    \put(0.05990291,0.01620532){\color[rgb]{0,0,0}\makebox(0,0)[lt]{\lineheight{1.25}\smash{\begin{tabular}[t]{l}0\end{tabular}}}}%
    \put(0.0988688,0.01620532){\color[rgb]{0,0,0}\makebox(0,0)[lt]{\lineheight{1.25}\smash{\begin{tabular}[t]{l}2\end{tabular}}}}%
    \put(0.13783471,0.01620532){\color[rgb]{0,0,0}\makebox(0,0)[lt]{\lineheight{1.25}\smash{\begin{tabular}[t]{l}4\end{tabular}}}}%
    \put(0.1768006,0.01620532){\color[rgb]{0,0,0}\makebox(0,0)[lt]{\lineheight{1.25}\smash{\begin{tabular}[t]{l}6\end{tabular}}}}%
    \put(0.21576649,0.01620532){\color[rgb]{0,0,0}\makebox(0,0)[lt]{\lineheight{1.25}\smash{\begin{tabular}[t]{l}8\end{tabular}}}}%
    \put(0.25228995,0.01620532){\color[rgb]{0,0,0}\makebox(0,0)[lt]{\lineheight{1.25}\smash{\begin{tabular}[t]{l}10\end{tabular}}}}%
    \put(0.02588535,0.04349668){\color[rgb]{0,0,0}\makebox(0,0)[lt]{\lineheight{1.25}\smash{\begin{tabular}[t]{l}10\end{tabular}}}}%
    \put(0,0){\includegraphics[width=\unitlength,page=2]{sims_pets.pdf}}%
    \put(0.03309153,0.06580919){\color[rgb]{0,0,0}\makebox(0,0)[lt]{\lineheight{1.25}\smash{\begin{tabular}[t]{l}0\end{tabular}}}}%
    \put(0.02588535,0.08812174){\color[rgb]{0,0,0}\makebox(0,0)[lt]{\lineheight{1.25}\smash{\begin{tabular}[t]{l}10\end{tabular}}}}%
    \put(0.28124284,0.01616063){\color[rgb]{0,0,0}\makebox(0,0)[lt]{\lineheight{1.25}\smash{\begin{tabular}[t]{l}0\end{tabular}}}}%
    \put(0.32015198,0.01616063){\color[rgb]{0,0,0}\makebox(0,0)[lt]{\lineheight{1.25}\smash{\begin{tabular}[t]{l}2\end{tabular}}}}%
    \put(0.35906114,0.01616063){\color[rgb]{0,0,0}\makebox(0,0)[lt]{\lineheight{1.25}\smash{\begin{tabular}[t]{l}4\end{tabular}}}}%
    \put(0.39797028,0.01616063){\color[rgb]{0,0,0}\makebox(0,0)[lt]{\lineheight{1.25}\smash{\begin{tabular}[t]{l}6\end{tabular}}}}%
    \put(0.43687944,0.01616063){\color[rgb]{0,0,0}\makebox(0,0)[lt]{\lineheight{1.25}\smash{\begin{tabular}[t]{l}8\end{tabular}}}}%
    \put(0.47334963,0.01616063){\color[rgb]{0,0,0}\makebox(0,0)[lt]{\lineheight{1.25}\smash{\begin{tabular}[t]{l}10\end{tabular}}}}%
    \put(0.49730909,0.03617998){\color[rgb]{0,0,0}\makebox(0,0)[lt]{\lineheight{1.25}\smash{\begin{tabular}[t]{l}-50\end{tabular}}}}%
    \put(0.00538338,0.15061001){\color[rgb]{0,0,0}\rotatebox{90}{\makebox(0,0)[lt]{\lineheight{1.25}\smash{\begin{tabular}[t]{l}$K$\end{tabular}}}}}%
    \put(0.00508594,0.06256545){\color[rgb]{0,0,0}\rotatebox{90}{\makebox(0,0)[lt]{\lineheight{1.25}\smash{\begin{tabular}[t]{l}$f_{ext}$\end{tabular}}}}}%
    \put(0.51191927,0.0885036){\color[rgb]{0,0,0}\makebox(0,0)[lt]{\lineheight{1.25}\smash{\begin{tabular}[t]{l}0\end{tabular}}}}%
    \put(0.50952738,0.21805818){\color[rgb]{0,0,0}\makebox(0,0)[lt]{\lineheight{1.25}\smash{\begin{tabular}[t]{l}0\end{tabular}}}}%
    \put(0.50972587,0.23799628){\color[rgb]{0,0,0}\makebox(0,0)[lt]{\lineheight{1.25}\smash{\begin{tabular}[t]{l}1\end{tabular}}}}%
    \put(0.50463727,0.19845783){\color[rgb]{0,0,0}\makebox(0,0)[lt]{\lineheight{1.25}\smash{\begin{tabular}[t]{l}-1\end{tabular}}}}%
    \put(0.75128744,0.05515183){\color[rgb]{0,0,0}\makebox(0,0)[lt]{\lineheight{1.25}\smash{\begin{tabular}[t]{l}-50\end{tabular}}}}%
    \put(0.76818285,0.09353313){\color[rgb]{0,0,0}\makebox(0,0)[lt]{\lineheight{1.25}\smash{\begin{tabular}[t]{l}0\end{tabular}}}}%
    \put(0.76818735,0.19776356){\color[rgb]{0,0,0}\makebox(0,0)[lt]{\lineheight{1.25}\smash{\begin{tabular}[t]{l}0\end{tabular}}}}%
    \put(0.75865359,0.2601291){\color[rgb]{0,0,0}\makebox(0,0)[lt]{\lineheight{1.25}\smash{\begin{tabular}[t]{l}10\end{tabular}}}}%
    \put(0.37457086,0.00121398){\color[rgb]{0,0,0}\makebox(0,0)[lt]{\lineheight{1.25}\smash{\begin{tabular}[t]{l}$t$\end{tabular}}}}%
    \put(0.15337635,0.00121398){\color[rgb]{0,0,0}\makebox(0,0)[lt]{\lineheight{1.25}\smash{\begin{tabular}[t]{l}$t$\end{tabular}}}}%
    \put(0.53455071,0.01616085){\color[rgb]{0,0,0}\makebox(0,0)[lt]{\lineheight{1.25}\smash{\begin{tabular}[t]{l}0\end{tabular}}}}%
    \put(0.57351661,0.01610239){\color[rgb]{0,0,0}\makebox(0,0)[lt]{\lineheight{1.25}\smash{\begin{tabular}[t]{l}2\end{tabular}}}}%
    \put(0.61248251,0.01615682){\color[rgb]{0,0,0}\makebox(0,0)[lt]{\lineheight{1.25}\smash{\begin{tabular}[t]{l}4\end{tabular}}}}%
    \put(0.6514484,0.01616085){\color[rgb]{0,0,0}\makebox(0,0)[lt]{\lineheight{1.25}\smash{\begin{tabular}[t]{l}6\end{tabular}}}}%
    \put(0.6904143,0.01616085){\color[rgb]{0,0,0}\makebox(0,0)[lt]{\lineheight{1.25}\smash{\begin{tabular}[t]{l}8\end{tabular}}}}%
    \put(0.72693774,0.01616085){\color[rgb]{0,0,0}\makebox(0,0)[lt]{\lineheight{1.25}\smash{\begin{tabular}[t]{l}10\end{tabular}}}}%
    \put(0.62704214,0.00121398){\color[rgb]{0,0,0}\makebox(0,0)[lt]{\lineheight{1.25}\smash{\begin{tabular}[t]{l}$t$\end{tabular}}}}%
    \put(0.79048058,0.01616085){\color[rgb]{0,0,0}\makebox(0,0)[lt]{\lineheight{1.25}\smash{\begin{tabular}[t]{l}0\end{tabular}}}}%
    \put(0.82944647,0.01610239){\color[rgb]{0,0,0}\makebox(0,0)[lt]{\lineheight{1.25}\smash{\begin{tabular}[t]{l}2\end{tabular}}}}%
    \put(0.86841237,0.01615682){\color[rgb]{0,0,0}\makebox(0,0)[lt]{\lineheight{1.25}\smash{\begin{tabular}[t]{l}4\end{tabular}}}}%
    \put(0.90737827,0.01616085){\color[rgb]{0,0,0}\makebox(0,0)[lt]{\lineheight{1.25}\smash{\begin{tabular}[t]{l}6\end{tabular}}}}%
    \put(0.94634416,0.01616085){\color[rgb]{0,0,0}\makebox(0,0)[lt]{\lineheight{1.25}\smash{\begin{tabular}[t]{l}8\end{tabular}}}}%
    \put(0.9828676,0.01616085){\color[rgb]{0,0,0}\makebox(0,0)[lt]{\lineheight{1.25}\smash{\begin{tabular}[t]{l}10\end{tabular}}}}%
    \put(0.88297201,0.00121398){\color[rgb]{0,0,0}\makebox(0,0)[lt]{\lineheight{1.25}\smash{\begin{tabular}[t]{l}$t$\end{tabular}}}}%
    \put(0,0){\includegraphics[width=\unitlength,page=3]{sims_pets.pdf}}%
    \put(0.94420445,0.24934192){\color[rgb]{0,0,0}\makebox(0,0)[lt]{\lineheight{1.25}\smash{\begin{tabular}[t]{l}$x$-axis\end{tabular}}}}%
    \put(0,0){\includegraphics[width=\unitlength,page=4]{sims_pets.pdf}}%
    \put(0.94420445,0.23026147){\color[rgb]{0,0,0}\makebox(0,0)[lt]{\lineheight{1.25}\smash{\begin{tabular}[t]{l}$y$-axis\end{tabular}}}}%
    \put(0,0){\includegraphics[width=\unitlength,page=5]{sims_pets.pdf}}%
    \put(0.94420445,0.20886027){\color[rgb]{0,0,0}\makebox(0,0)[lt]{\lineheight{1.25}\smash{\begin{tabular}[t]{l}$z$-axis\end{tabular}}}}%
    \put(0,0){\includegraphics[width=\unitlength,page=6]{sims_pets.pdf}}%
    \put(0.63014974,0.27776115){\color[rgb]{0,0,0}\makebox(0,0)[lt]{\lineheight{1.25}\smash{\begin{tabular}[t]{l}(c)\end{tabular}}}}%
    \put(0,0){\includegraphics[width=\unitlength,page=7]{sims_pets.pdf}}%
    \put(0.88768614,0.27729806){\color[rgb]{0,0,0}\makebox(0,0)[lt]{\lineheight{1.25}\smash{\begin{tabular}[t]{l}(d)\end{tabular}}}}%
  \end{picture}%
\endgroup%

%% file: figures/reward_transfer.pdf_tex
\begingroup%
  \makeatletter%
  \providecommand\color[2][]{%
    \errmessage{(Inkscape) Color is used for the text in Inkscape, but the package 'color.sty' is not loaded}%
    \renewcommand\color[2][]{}%
  }%
  \providecommand\transparent[1]{%
    \errmessage{(Inkscape) Transparency is used (non-zero) for the text in Inkscape, but the package 'transparent.sty' is not loaded}%
    \renewcommand\transparent[1]{}%
  }%
  \providecommand\rotatebox[2]{#2}%
  \newcommand*\fsize{\dimexpr\f@size pt\relax}%
  \newcommand*\lineheight[1]{\fontsize{\fsize}{#1\fsize}\selectfont}%
  \ifx\svgwidth\undefined%
    \setlength{\unitlength}{301.02330625bp}%
    \ifx\svgscale\undefined%
      \relax%
    \else%
      \setlength{\unitlength}{\unitlength * \real{\svgscale}}%
    \fi%
  \else%
    \setlength{\unitlength}{\svgwidth}%
  \fi%
  \global\let\svgwidth\undefined%
  \global\let\svgscale\undefined%
  \makeatother%
  \begin{picture}(1,0.29509213)%
    \lineheight{1}%
    \setlength\tabcolsep{0pt}%
    \put(0,0){\includegraphics[width=\unitlength,page=1]{reward_transfer.pdf}}%
    \put(0.02380944,0.24201019){\color[rgb]{0,0,0}\makebox(0,0)[lt]{\lineheight{1.25}\smash{\begin{tabular}[t]{l}1.0\end{tabular}}}}%
    \put(0.02448126,0.13764289){\color[rgb]{0,0,0}\makebox(0,0)[lt]{\lineheight{1.25}\smash{\begin{tabular}[t]{l}0.5\end{tabular}}}}%
    \put(0.02426405,0.03431434){\color[rgb]{0,0,0}\makebox(0,0)[lt]{\lineheight{1.25}\smash{\begin{tabular}[t]{l}0.0\end{tabular}}}}%
    \put(0.01269075,0.09048261){\rotatebox{90}{\makebox(0,0)[lt]{\lineheight{1.25}\smash{\begin{tabular}[t]{l}Reward\end{tabular}}}}}%
    \put(0.10199953,0.00030428){\makebox(0,0)[lt]{\lineheight{1.25}\smash{\begin{tabular}[t]{l}Task a\end{tabular}}}}%
    \put(0.26539075,0.00030428){\makebox(0,0)[lt]{\lineheight{1.25}\smash{\begin{tabular}[t]{l}Task b\end{tabular}}}}%
    \put(0.43003792,0.00030428){\makebox(0,0)[lt]{\lineheight{1.25}\smash{\begin{tabular}[t]{l}Task c\end{tabular}}}}%
    \put(0.59649512,0.00030428){\makebox(0,0)[lt]{\lineheight{1.25}\smash{\begin{tabular}[t]{l}Task b\end{tabular}}}}%
    \put(0.87743693,0.00030428){\makebox(0,0)[lt]{\lineheight{1.25}\smash{\begin{tabular}[t]{l}Task c\end{tabular}}}}%
    \put(0,0){\includegraphics[width=\unitlength,page=2]{reward_transfer.pdf}}%
    \put(0.71467232,0.21402807){\makebox(0,0)[lt]{\lineheight{1.25}\smash{\begin{tabular}[t]{l}\ac{mpvic}\end{tabular}}}}%
    \put(0.71519286,0.17338811){\makebox(0,0)[lt]{\lineheight{1.25}\smash{\begin{tabular}[t]{l}\ac{rl}\end{tabular}}}}%
    \put(0.71519286,0.13275247){\makebox(0,0)[lt]{\lineheight{1.25}\smash{\begin{tabular}[t]{l}PETS\end{tabular}}}}%
    \put(0,0){\includegraphics[width=\unitlength,page=3]{reward_transfer.pdf}}%
    \put(0.21652177,0.27876605){\makebox(0,0)[lt]{\lineheight{1.25}\smash{\begin{tabular}[t]{l}\textit{performance}\end{tabular}}}}%
    \put(0.68870206,0.27876605){\makebox(0,0)[lt]{\lineheight{1.25}\smash{\begin{tabular}[t]{l}\textit{transferability}\end{tabular}}}}%
  \end{picture}%
\endgroup%

%% file: figures/experiments.pdf_tex
\begingroup%
  \makeatletter%
  \providecommand\color[2][]{%
    \errmessage{(Inkscape) Color is used for the text in Inkscape, but the package 'color.sty' is not loaded}%
    \renewcommand\color[2][]{}%
  }%
  \providecommand\transparent[1]{%
    \errmessage{(Inkscape) Transparency is used (non-zero) for the text in Inkscape, but the package 'transparent.sty' is not loaded}%
    \renewcommand\transparent[1]{}%
  }%
  \providecommand\rotatebox[2]{#2}%
  \newcommand*\fsize{\dimexpr\f@size pt\relax}%
  \newcommand*\lineheight[1]{\fontsize{\fsize}{#1\fsize}\selectfont}%
  \ifx\svgwidth\undefined%
    \setlength{\unitlength}{323.87038171bp}%
    \ifx\svgscale\undefined%
      \relax%
    \else%
      \setlength{\unitlength}{\unitlength * \real{\svgscale}}%
    \fi%
  \else%
    \setlength{\unitlength}{\svgwidth}%
  \fi%
  \global\let\svgwidth\undefined%
  \global\let\svgscale\undefined%
  \makeatother%
  \begin{picture}(1,0.69207805)%
    \lineheight{1}%
    \setlength\tabcolsep{0pt}%
    \put(0,0){\includegraphics[width=\unitlength,page=1]{experiments.pdf}}%
    \put(0.01496684,0.57047989){\color[rgb]{0,0,0}\rotatebox{90}{\makebox(0,0)[lt]{\lineheight{1.25}\smash{\begin{tabular}[t]{l}$\Delta_{pos}$\end{tabular}}}}}%
    \put(0.0186707,0.38028982){\color[rgb]{0,0,0}\rotatebox{90}{\makebox(0,0)[lt]{\lineheight{1.25}\smash{\begin{tabular}[t]{l}$K$\end{tabular}}}}}%
    \put(0.01810112,0.13873437){\color[rgb]{0,0,0}\rotatebox{90}{\makebox(0,0)[lt]{\lineheight{1.25}\smash{\begin{tabular}[t]{l}$f_{ext}$\end{tabular}}}}}%
    \put(0.04430915,0.66489927){\color[rgb]{0,0,0}\makebox(0,0)[lt]{\lineheight{1.25}\smash{\begin{tabular}[t]{l}0\end{tabular}}}}%
    \put(0.03019062,0.50951748){\color[rgb]{0,0,0}\makebox(0,0)[lt]{\lineheight{1.25}\smash{\begin{tabular}[t]{l}-15\end{tabular}}}}%
    \put(0.02666641,0.45700755){\color[rgb]{0,0,0}\makebox(0,0)[lt]{\lineheight{1.25}\smash{\begin{tabular}[t]{l}200\end{tabular}}}}%
    \put(0.05877362,0.29920564){\color[rgb]{0,0,0}\makebox(0,0)[lt]{\lineheight{1.25}\smash{\begin{tabular}[t]{l}0\end{tabular}}}}%
    \put(0.03496847,0.25894955){\color[rgb]{0,0,0}\makebox(0,0)[lt]{\lineheight{1.25}\smash{\begin{tabular}[t]{l}10\end{tabular}}}}%
    \put(0.0300362,0.08308042){\color[rgb]{0,0,0}\makebox(0,0)[lt]{\lineheight{1.25}\smash{\begin{tabular}[t]{l}-20\end{tabular}}}}%
    \put(0.55706754,0.54726943){\color[rgb]{0,0,0}\makebox(0,0)[lt]{\lineheight{1.25}\smash{\begin{tabular}[t]{l}0\end{tabular}}}}%
    \put(0.55706754,0.30676098){\color[rgb]{0,0,0}\makebox(0,0)[lt]{\lineheight{1.25}\smash{\begin{tabular}[t]{l}0\end{tabular}}}}%
    \put(0.55706754,0.24207036){\color[rgb]{0,0,0}\makebox(0,0)[lt]{\lineheight{1.25}\smash{\begin{tabular}[t]{l}0\end{tabular}}}}%
    \put(0.09950519,0.04753514){\color[rgb]{0,0,0}\makebox(0,0)[lt]{\lineheight{1.25}\smash{\begin{tabular}[t]{l}0\end{tabular}}}}%
    \put(0.55731532,0.64696256){\color[rgb]{0,0,0}\makebox(0,0)[lt]{\lineheight{1.25}\smash{\begin{tabular}[t]{l}2\end{tabular}}}}%
    \put(0.51379137,0.44885336){\color[rgb]{0,0,0}\makebox(0,0)[lt]{\lineheight{1.25}\smash{\begin{tabular}[t]{l}400\end{tabular}}}}%
    \put(0.52621686,0.0848674){\color[rgb]{0,0,0}\makebox(0,0)[lt]{\lineheight{1.25}\smash{\begin{tabular}[t]{l}-10\end{tabular}}}}%
    \put(0,0){\includegraphics[width=\unitlength,page=2]{experiments.pdf}}%
    \put(0.36008724,0.61673873){\color[rgb]{0,0,0}\makebox(0,0)[lt]{\lineheight{1.25}\smash{\begin{tabular}[t]{l}$x$-axis\end{tabular}}}}%
    \put(0,0){\includegraphics[width=\unitlength,page=3]{experiments.pdf}}%
    \put(0.36008724,0.5771249){\color[rgb]{0,0,0}\makebox(0,0)[lt]{\lineheight{1.25}\smash{\begin{tabular}[t]{l}$y$-axis\end{tabular}}}}%
    \put(0,0){\includegraphics[width=\unitlength,page=4]{experiments.pdf}}%
    \put(0.36008724,0.53114388){\color[rgb]{0,0,0}\makebox(0,0)[lt]{\lineheight{1.25}\smash{\begin{tabular}[t]{l}$z$-axis\end{tabular}}}}%
    \put(0.17552715,0.04752411){\color[rgb]{0,0,0}\makebox(0,0)[lt]{\lineheight{1.25}\smash{\begin{tabular}[t]{l}1\end{tabular}}}}%
    \put(0.25133203,0.0473748){\color[rgb]{0,0,0}\makebox(0,0)[lt]{\lineheight{1.25}\smash{\begin{tabular}[t]{l}2\end{tabular}}}}%
    \put(0.32553844,0.04753514){\color[rgb]{0,0,0}\makebox(0,0)[lt]{\lineheight{1.25}\smash{\begin{tabular}[t]{l}3\end{tabular}}}}%
    \put(0.39971717,0.04752411){\color[rgb]{0,0,0}\makebox(0,0)[lt]{\lineheight{1.25}\smash{\begin{tabular}[t]{l}4\end{tabular}}}}%
    \put(0.47573915,0.04768446){\color[rgb]{0,0,0}\makebox(0,0)[lt]{\lineheight{1.25}\smash{\begin{tabular}[t]{l}5\end{tabular}}}}%
    \put(0.59306245,0.04753514){\color[rgb]{0,0,0}\makebox(0,0)[lt]{\lineheight{1.25}\smash{\begin{tabular}[t]{l}0\end{tabular}}}}%
    \put(0.71062133,0.04753514){\color[rgb]{0,0,0}\makebox(0,0)[lt]{\lineheight{1.25}\smash{\begin{tabular}[t]{l}10\end{tabular}}}}%
    \put(0.83680891,0.04753514){\color[rgb]{0,0,0}\makebox(0,0)[lt]{\lineheight{1.25}\smash{\begin{tabular}[t]{l}20\end{tabular}}}}%
    \put(0.96085731,0.04753514){\color[rgb]{0,0,0}\makebox(0,0)[lt]{\lineheight{1.25}\smash{\begin{tabular}[t]{l}30\end{tabular}}}}%
    \put(0.78604197,0.00333063){\color[rgb]{0,0,0}\makebox(0,0)[lt]{\lineheight{1.25}\smash{\begin{tabular}[t]{l}$t$\end{tabular}}}}%
    \put(0.29303889,0.00333063){\color[rgb]{0,0,0}\makebox(0,0)[lt]{\lineheight{1.25}\smash{\begin{tabular}[t]{l}$t$\end{tabular}}}}%
  \end{picture}%
\endgroup%